\documentclass[11pt]{article}

\usepackage[preprint]{acl}

\usepackage{times}
\usepackage{latexsym}
\usepackage[T1]{fontenc}
\usepackage[utf8]{inputenc}
\usepackage{microtype}
\usepackage{inconsolata}

\usepackage{graphicx}
\usepackage{subcaption}
\usepackage{booktabs}
\usepackage{multirow}
\usepackage{amsmath}
\usepackage{amssymb}
\usepackage{mathtools}
\usepackage{amsthm}
\usepackage[capitalize,noabbrev]{cleveref}
\usepackage{pifont}

\theoremstyle{plain}

\theoremstyle{definition}

\theoremstyle{remark}

\usepackage[most]{tcolorbox}
\usepackage{upquote}

\definecolor{PromptBack}{HTML}{F8FAFC}
\definecolor{PromptFrame}{HTML}{6B7280}
\definecolor{PromptTitleBack}{HTML}{EEF2F7}

\newtcblisting{PromptBox}[2][]{%
  enhanced,
  breakable,
  width=\linewidth,
  colback=PromptBack,
  colframe=PromptFrame,
  coltitle=black,
  colbacktitle=PromptTitleBack,
  boxrule=0.45pt,
  arc=1mm,
  left=1.2mm,
  right=1.2mm,
  top=1mm,
  bottom=1mm,
  before skip=0.75\baselineskip,
  after skip=0.75\baselineskip,
  title={#2},
  fonttitle=\bfseries\footnotesize,
  listing only,
  listing options={
    basicstyle=\ttfamily\scriptsize,
    columns=fullflexible,
    keepspaces=true,
    breaklines=true,
    breakatwhitespace=false,
    showstringspaces=false,
    upquote=true,
    aboveskip=0pt,
    belowskip=0pt
  },
  #1
}

\title{Unstable Features, Reproducible Subspaces: Understanding Seed Dependence in Sparse Autoencoders}

\author{
  Gleb Gerasimov
  \And
  Timofei Rusalev
  \And
  Nikita Balagansky
  \And
  Daniil Laptev
  \AND
  Vadim Kurochkin 
  \And
  Daniil Gavrilov
  \AND
  \normalfont
  T-Tech
}

\begin{document}
\maketitle
\begin{abstract}

Sparse autoencoders (SAEs) are widely used to interpret neural network representations, but their utility depends on whether the learned features are reproducible across training runs. We study this question through \emph{feature stability}: for each SAE feature, we estimate the probability that a similar feature reappears in an independently trained SAE. This yields a scalable per-feature signal that separates stable from unstable features. In a large-scale study across seeds, models, layers, dictionary sizes, and SAE variants, we find a pronounced functional asymmetry: stable features carry most of the reconstruction- and prediction-relevant signal, while unstable features have weak marginal impact and are dominated by low-frequency surface-form triggers in both activation statistics and automatic explanations. Geometrically, unstable features are individually non-reproducible but concentrate in reproducible lower-rank subspaces, suggesting that seed dependence often reflects basis ambiguity within a shared region of activation space rather than pure noise. A controlled synthetic model makes this mechanism explicit, showing that low-rank ground-truth features can be recovered at the subspace level while remaining non-identifiable as individual SAE latents across seeds. Finally, by pooling unique cross-seed features, we construct more stable SAEs while preserving explained variance in this setting. Together, these results show that unstable features are not merely failed or noisy latents: they have weak individual functional impact, but reflect reproducible low-dimensional structure that standard SAEs resolve differently across seeds.
\end{abstract}

\section{Introduction}

Sparse autoencoders (SAEs) are a central tool in mechanistic interpretability because they aim to decompose model activations into sparse, human-interpretable features \citep{bricken2023towards,cunningham2023sparse,gao2025scaling}. A basic question, however, is whether these features are reproducible: prior work has shown that SAEs trained on the same activations but with different random seeds can learn substantially different features \citep{paulo2025samedata,leask2025canonical}.

 This raises a finer-grained question: when an individual feature fails to reappear, is the underlying direction absent from the new dictionary, or has the SAE learned a different basis for the same region of activation space? This distinction matters because automatic interpretation can assign plausible explanations even in random or non-canonical settings \citep{heap2025automated,bhalla2026manifolds}, so an interpretable-looking feature need not be a reproducible unit across runs.

We study this question through \emph{feature stability}: the probability that a feature reappears in an independently trained SAE under a cosine-similarity matching rule. Using this per-feature signal, we compare stable and unstable features by their functional impact, token structure, automatic explanations, and decoder-space geometry.

Our contributions are as follows.

\begin{itemize}

\item \textbf{Functional asymmetry of stable and unstable features.}
Stable features carry most of the reconstruction- and prediction-relevant signal, whereas unstable features have weak marginal impact and are dominated by low-frequency surface-form triggers in both activation statistics and automatic explanations.

\item \textbf{Subspace recovery without feature identifiability.}
Decoder-space analysis shows that unstable features are individually non-reproducible but collectively span reproducible lower-rank subspaces. In a controlled synthetic setting, low-rank ground-truth features are likewise recovered at the subspace level while failing to align with individual SAE latents.

\item \textbf{No stability--EV trade-off in feature-pool construction.}
Using a deduplicated pool of unique features from multiple runs, we construct SAEs from stable cross-seed features and find that higher explained variance coincides with higher mean feature probability after tuning, suggesting that stability and reconstruction quality need not trade off in this construction setting.

\end{itemize}

\section{Related Work}

\textbf{Feature stability.} \citet{paulo2025samedata} showed that SAEs trained on the same model and data, differing only in random seed, can learn substantially different feature sets. \citet{leask2025canonical} argued that SAE latents are not canonical units of analysis, showing that features vary with dictionary size and can be merged across dictionaries. \citet{fel2025archetypal} mitigate instability by anchoring dictionary atoms to data-derived archetypes. \citet{gadgil2025ensembling} propose ensembling independently initialized SAEs via naive bagging and boosting. Our construction of an SAE from unique cross-seed features follows the same bagging intuition, but deduplicates the pooled feature set and selects a fixed-size subset of high-stability features rather than simply concatenating all latents from all runs. \citet{chen2025taming} improve consistency by assigning features to activation-frequency groups and adapting encoder biases to enforce group-specific target frequencies. \citet{cho2025faithfulsae} argue that external training data can be partially OOD for the base model and report that training on model-generated data improves cross-seed stability. \citet{wang2025orderedness} impose an ordered latent structure in which increasingly large prefixes of the feature set reconstruct the activation, reducing permutation non-identifiability and improving consistency. \citet{heap2025automated} showed that SAEs trained on randomly initialized transformers can still yield plausible automatic interpretations, motivating random-model controls for SAE interpretability. \citet{bhalla2026manifolds} give a complementary geometric account, arguing that multidimensional concepts can admit multiple valid SAE bases, making seed-dependent decompositions expected.

In this work, we estimate per-feature reappearance probabilities to study instability through seed-dependent basis choices within reproducible lower-rank decoder subspaces, probing this view with a controlled low-rank synthetic model and comparing stable and unstable features by reconstruction impact, downstream effects, and automatic explanations.

\textbf{Finding similar features across SAEs.} \citet{balcells2024evolution} track how individual SAE features evolve between adjacent layers, while \citet{balagansky2025permutability} introduce a data-free method for matching SAE features across layers; \citet{laptev2025featureflow} extend this line by constructing cross-layer feature-flow graphs. \citet{wang2025universality} identify corresponding SAE features across different language-model architectures. \citet{anonymous2026crossseed} bridge the stability and matching literatures by benchmarking cross-seed SAE correspondence methods, including cosine and optimal-transport matchers, and evaluating them functionally via ablation effects and substitution tests.

\section{Preliminaries}

\subsection{Sparse Autoencoders}

A sparse autoencoder (SAE) is designed to represent hidden states as a sparse linear combination of feature embeddings \citep{bricken2023towards, cunningham2023sparse}; each coefficient is treated as an activation magnitude, or importance of this feature, which are assumed to be non-negative.

Architecturally, an SAE has the following form:
\begin{equation}
\begin{gathered}
    \boldsymbol{z} = \sigma (\boldsymbol{W}_{\text{enc}} \boldsymbol{h} + \boldsymbol{b}_{\text{enc}}) \in \mathbb{R}^F, \\
    \hat{\boldsymbol{h}} = \boldsymbol{W}_{\text{dec}} \boldsymbol{z} + \boldsymbol{b}_{\text{dec}} \in \mathbb{R}^d,
\end{gathered}
\end{equation}
where $\boldsymbol{h}$ is a hidden state vector, $\boldsymbol{z}$ is a vector of feature activation magnitudes, and $\sigma$ is a nonlinear activation function that induces sparsity and non-negativity. SAEs are trained on reconstruction loss plus optional regularization loss that controls sparsity level scaled by coefficient $\alpha \ge 0$:
\begin{equation}
    \mathcal{L} = \| \boldsymbol{h} - \hat{\boldsymbol{h}} \|_2^2 + \alpha \mathcal{L}_{\text{reg}}(\boldsymbol{z}).
\end{equation}
We consider five SAE variants used in our experiments: Vanilla ReLU+$\ell_1$ SAEs, TopK SAEs, BatchTopK SAEs, HierarchicalTopK SAEs, and JumpReLU SAEs
\citep{cunningham2023sparse,gao2025scaling,bussmann2024batchtopksparseautoencoders,balagansky-etal-2025-train,rajamanoharan2024jumpingahead}.
TopK and BatchTopK impose fixed sparsity by retaining only the largest activations, HierarchicalTopK trains one SAE across multiple sparsity budgets, and JumpReLU uses learned per-feature thresholds.
For feature \(i\) in SAE \(s\), we write \(\boldsymbol{f}^{(s)}_i\) for the feature, \(\boldsymbol{z}^{(s)}_i\) for its activation, and \(\boldsymbol{e}^{(s)}_i\in\mathbb{R}^d\) for its decoder embedding; decoder columns are unit-normalized after training (Appendix~\ref{sec:appendix:training_details}).

\subsection{Feature Matching and Stability}

Two SAEs trained on similar hidden state distribution (e.g., on the same layer) are expected to converge towards similar representations of the hidden states and therefore similar feature representations. Since SAEs are invariant to feature ordering and two different initializations might produce same dictionaries, but with different indices \citep{balagansky2025permutability}, a linear assignment problem is usually solved \citep{paulo2025samedata, balagansky2025permutability, fel2025archetypal}: given two dictionaries $\boldsymbol{D}_{1}$ and $\boldsymbol{D}_{2}$, we seek for a permutation matrix $\boldsymbol{\Pi}$ so that discrepancy between dictionaries $\boldsymbol{D}_1$ and $\boldsymbol{\Pi} \boldsymbol{D}_2$ is minimized; this gives a one-to-one correspondence. The quality of resulting matching is a measure of SAE stability: if it is low, then features are mostly different. This is primary method for previous works on the problem we investigate in this paper.

In this work we adopt a many-to-one alternative: for each element from $\boldsymbol{D}_1$, we find the most similar element from $\boldsymbol{D}_2$, measured by maximum cosine similarity. This approach is much less computationally expensive and allows to compute stability of individual features, helping us understand how their individual properties correspond to the problem of instability. As a robustness check, replacing this argmax-cosine rule with one-to-one Hungarian matching yields nearly identical matched feature sets (IoU $=0.978\pm0.001$).

\section{Methodology}
\label{sec:stability-metric}

We study how reliably individual SAE features reappear across random initializations.
Throughout, we represent each feature by its decoder vector, and $\ell_2$-normalize all decoder columns so that cosine similarity reduces to a dot product.

\textbf{Feature matching.}
Given two SAEs $A$ and $B$ with decoder columns $\{\boldsymbol{e}^{(A)}_i\}_{i=1}^F$ and $\{\boldsymbol{e}^{(B)}_j\}_{j=1}^F$, we say that features $i$ and $j$ \emph{match} if $\cos\!\big(\boldsymbol{e}^{(A)}_i,\boldsymbol{e}^{(B)}_j\big)\ \ge\ \theta,$
following \cite{leask2025canonical}, $\theta=0.7$ unless stated otherwise.

\textbf{Reappearance probability.}
We train $N\!+\!1$ SAEs on identical data and hyperparameters with different random seeds, and choose one run as an anchor ($k=0$).
For each anchor feature embedding $\boldsymbol{e}^{(0)}_i$, we count how many of the other $N$ SAEs contain a feature with cosine similarity at least $\theta$:
\begin{equation}
\label{eq:count_reappear}
X_{0,i}
=
\sum_{k=1}^{N}
\boldsymbol{1}\!\left\{
\max_{j\in\{1,\dots,F\}}
\cos\!\big(\boldsymbol{e}^{(0)}_i,\boldsymbol{e}^{(k)}_j\big)
\ge \theta
\right\}.
\end{equation}
Let
\[
p_i := \Pr\!\left(\max_{j} \cos(\boldsymbol{e}_i^{(0)}, \boldsymbol{e}_j^{(k)}) \ge \theta\right), \qquad k\neq 0.
\]

Under independent seeds, \(X_{0,i}\) is binomial with parameters \(N\) and \(p_i\), and we estimate
\begin{equation}
\label{eq:phat_reappear}
\hat{p}_i=\frac{X_{0,i}}{N}.
\end{equation}

Across features, we use the empirical CDF $\hat{\mathbf{F}}(p)=\frac{1}{F}\sum_{i=1}^{F}\boldsymbol{1}\{\hat{p}_i\le p\}$, which converges uniformly to $\mathbf{F}(p)$ (Glivenko--Cantelli; \citep{glivenko1933,cantelli1933}).

\textbf{Endpoint stability.}
We focus on \emph{endpoint} behavior: features that reappear in almost none or almost all runs.
Fixing $\varepsilon=0.05$, we call a feature \emph{unstable} if $\hat{p}(\boldsymbol{f}^{(0)}_i)\leq\varepsilon$ and \emph{stable} if $\hat{p}(\boldsymbol{f}^{(0)}_i)\geq1-\varepsilon$ and use notations
\[
\begin{aligned}
\mathcal{U}^{(s)}_\varepsilon
&= \{i:\hat{p}(\boldsymbol{f}^{(s)}_i)\le \varepsilon\},\\
\mathcal{S}^{(s)}_\varepsilon
&= \{i:\hat{p}(\boldsymbol{f}^{(s)}_i)\ge 1-\varepsilon\}.
\end{aligned}
\]
Our main task is therefore distilled into finding $\mathbf{F}(\varepsilon)$ and $\mathbf{F}(1 - \varepsilon)$, via empirical distribution function.

\textbf{Unique feature pool.}
For the construction experiment, we pool decoder features from several independently trained SAEs and greedily deduplicate them under the same cosine threshold \(\theta=0.7\), producing a unique feature pool \(\mathbb{U}\) used to initialize new SAEs (Section~\ref{sec:sae_sampling}; formal definition in Appendix~\ref{app:ev-vs-mp}).


\section{Quantitative Analysis}
\label{sec:quant}

\begin{figure}[t]
\centering
\includegraphics[width=\linewidth]{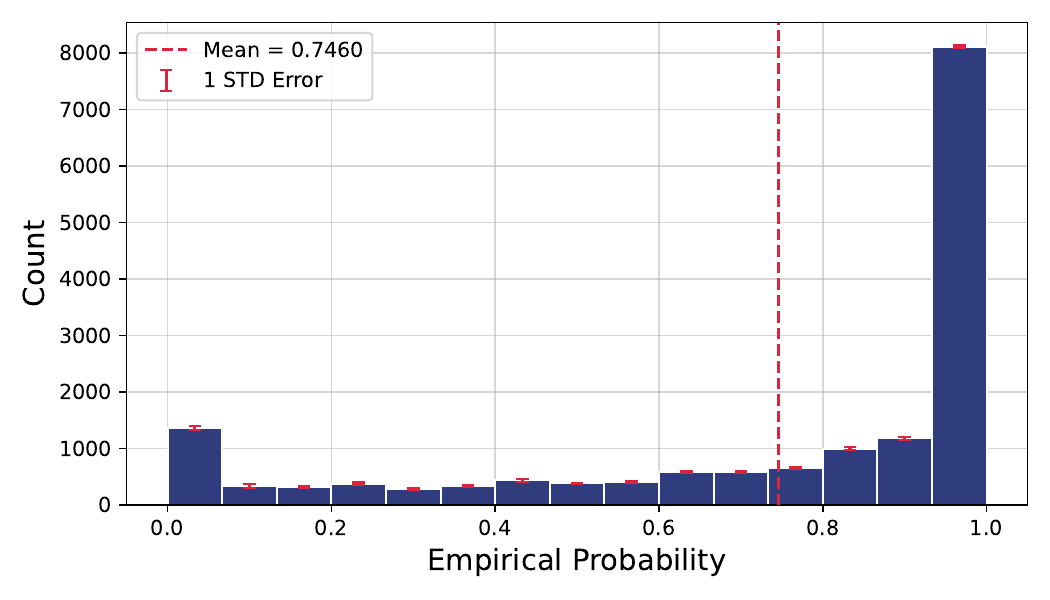}
\caption{
Feature reappearance across seeds in the main setup (See Section~\ref{sec:quant}).
Empirical distribution of reappearance rates $\hat p$ (Eq.~\ref{eq:phat_reappear}) for anchor features;
the vertical dashed line marks the mean, and error bars indicate variability across 5 anchor choices.
}
\label{fig:stability-main}
\label{fig:prob-hist}
\end{figure}

\paragraph{Experimental Setup.}
Unless stated otherwise, we train $96$ TopK SAEs (so $N=95$ comparisons per anchor) on GPT-2 residual-stream activations at layer $7$, with TopK $=64$ and dictionary size $F=2^{14}$.
Figure~\ref{fig:stability-main} summarizes the resulting reappearance rates $\hat p$: many features recur in nearly every run, a smaller separated group recurs in almost none, and relatively few lie in the middle (mean $\hat p\approx0.75$).
Sensitivity to the cosine threshold is reported in Appendix~\ref{app:threshold-sensitivity}; broader SAE-family comparisons are deferred to Section~\ref{sec:other-setups}.

\begin{figure}[t]
\vspace{-1.5mm}
\centering
\includegraphics[width=\linewidth]{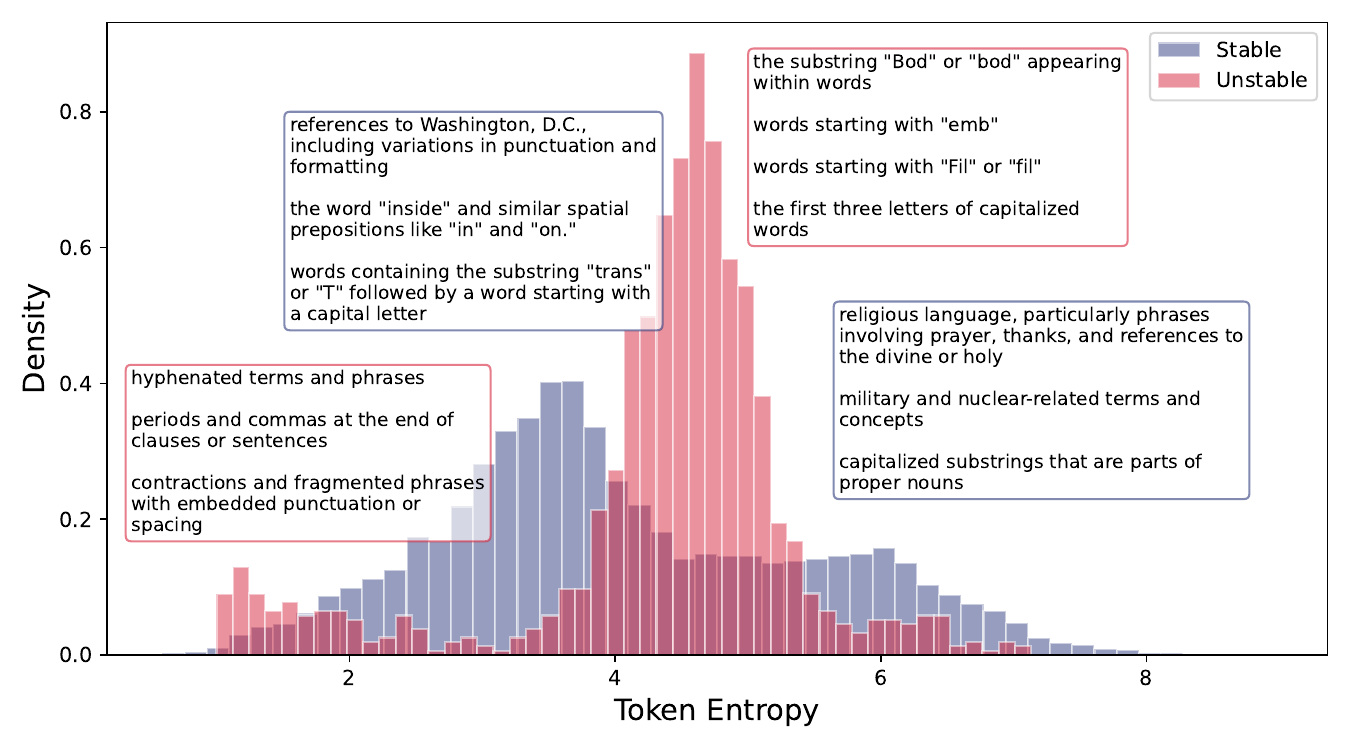}
\vspace{-1mm}
\caption{
\textbf{Token diversity for stable vs.\ unstable features.}
Token entropy $H_i$ with representative feature interpretations.
}
\label{fig:act-entropy}
\vspace{-2mm}
\end{figure}

\begin{figure*}[t]
\vspace{-1.5mm}
\centering
\includegraphics[width=\textwidth]{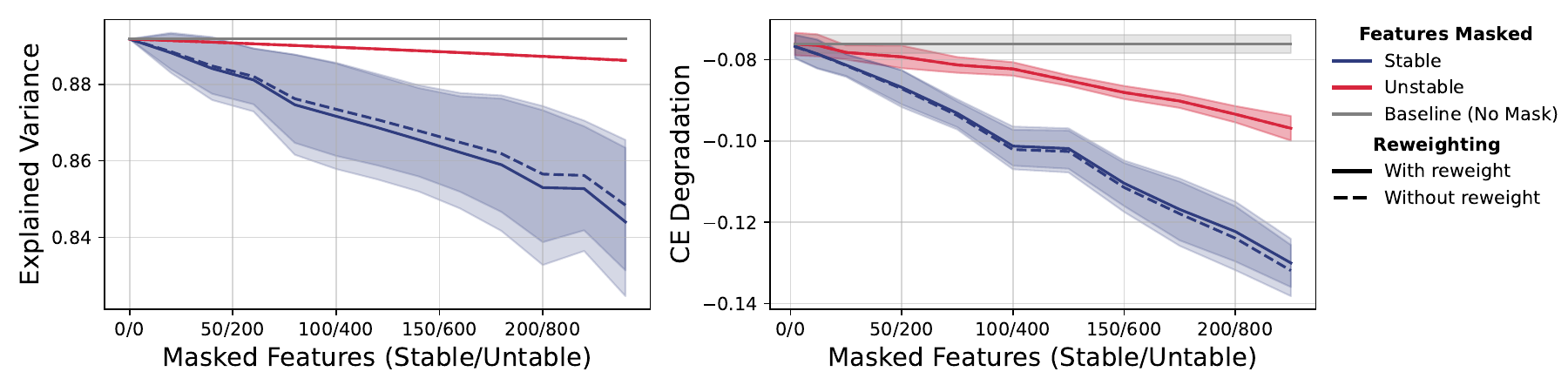}
\vspace{-2mm}
\caption{
\textbf{Impact of masking stable vs.\ unstable features on reconstruction and next-token loss.}
\textbf{Left}: explained variance (EV) under feature masking.
\textbf{Right}: change in next-token loss under activation patching with masked-feature reconstructions.
We mask $N$ stable features and $4N$ unstable features to approximately match expected active mass.
Solid vs.\ dashed curves correspond to using reweighting vs.\ not using reweighting; full protocol details are deferred to
Appendix~\ref{app:masking-protocol}.
}
\label{fig:ev-ce}
\vspace{-2mm}
\end{figure*}

We now use stability to compare \emph{what unstable vs.\ stable features look like} and \emph{how much they matter}.
Concretely, we show that: (i) unstable features activate less frequently and (on average) with smaller magnitude tails
(as quantified in Section~\ref{subsec:activation-stats} and shown in Appendix~\ref{app:usage-stats}, Figure~\ref{fig:mean-act-freq}); (ii) unstable features concentrate on lower-level lexical triggers with lower token
diversity (Figure~\ref{fig:act-entropy}); and (iii) even after aggressively masking them, unstable features have
much smaller impact on reconstruction and next-token loss than stable features under a frequency-matched protocol
(Figure~\ref{fig:ev-ce}).

All statistics below are computed on a held-out token collection; full evaluation and sampling details are deferred to
Appendix~\ref{app:quant-details}.


\subsection{Activation Statistics and Token Structure}
\label{subsec:activation-stats}

\textbf{Activation frequency and magnitude.}
We summarize feature usage with activation frequency $\omega_i$ (fraction of evaluated token positions where feature $i$
activates) and conditional mean magnitude $\mu_i$ (mean activation value conditioned on activation); formal definitions are
in Appendix~\ref{app:usage-stats}.
Appendix~\ref{app:usage-stats} (Figure~\ref{fig:mean-act-freq}) shows that unstable features activate less frequently on average:
their mean frequency is $\overline{\omega}_{\mathcal{U}_\varepsilon}\approx 0.0018$ (0.18\%),
compared to $\overline{\omega}_{\mathcal{S}_\varepsilon}\approx 0.0044$ (0.44\%) for stable features.
Stable features also exhibit a heavier high-magnitude tail in $\mu_i$.
(We defer dataset, batching, and exact $N_{\mathrm{tok}}$ details to Appendix~\ref{app:quant-eval}.)

\textbf{Token entropy (token diversity).}
We measure token diversity for feature \(i\) by the entropy \(H_i\) of the empirical distribution over vocabulary IDs at positions where \(z_{n,i}>0\) (definition in Appendix~\ref{app:token-entropy}). Low entropy indicates activation on a small token set; high entropy indicates broader lexical support.

Both stable and unstable features exhibit two dominant entropy regimes, but with different content. Unstable features range from punctuation/formatting triggers to short subword fragments and brittle substrings, while stable features range from single words or tight synonym clusters to higher-level concepts with many lexical realizations. Representative interpretations are shown in Figure~\ref{fig:act-entropy}.

\textbf{Automatic interpretability.}
SAEBench auto-interpretation mirrors this split (Appendix~\ref{app:qual-details}): stable features have higher detection scores (Figure~\ref{fig:autointerp}), with \(4.5\times\) more perfect-score features than unstable ones. Their explanations also differ systematically: unstable features are more often described by surface-form triggers (substrings, capitalization, punctuation), whereas stable features more often describe phrases, syntactic roles, constructions, and broader semantic groupings (Table~\ref{tab:interp-examples}). Keyword frequencies confirm this pattern: \texttt{substring} appears in \(38.9\%\) of unstable vs.\ \(11.3\%\) of stable explanations, while \texttt{phrase} rises from \(4.1\%\) to \(32.0\%\). GPT-5 predicts stable vs.\ unstable from explanation text alone with \(0.88\) accuracy (Appendix~\ref{app:gpt5-prompts}).


\subsection{Impact on Reconstruction and Next-Token Loss}
\label{subsec:functional-importance}

\textbf{Frequency-matched masking protocol.}
To control for the lower activation frequency of unstable features (Section~\ref{subsec:activation-stats}), for each $N$
we mask $N$ features sampled uniformly at random from $\mathcal{S}_\varepsilon$ and $4N$ features sampled uniformly at random
from $\mathcal{U}_\varepsilon$; full sampling, uncertainty, and reweighting details are deferred to
Appendix~\ref{app:masking-protocol}.

\textbf{Explained variance (EV).}
We measure how much reconstruction quality degrades after masking selected features by evaluating explained variance on the
resulting SAE reconstructions (formal definition in Appendix~\ref{app:masking-protocol}).
Figure~\ref{fig:ev-ce} (left) shows that masking unstable features changes EV only slightly even when masking $4N$ of them,
whereas masking far fewer stable features yields a substantially larger EV drop. This indicates that stable features account
for most of the reconstruction-relevant variance.

\textbf{Next-token loss under activation patching.}
To quantify downstream impact on the base model, we patch the residual stream at the SAE training location with the
masked-feature reconstruction and measure the resulting change in next-token cross-entropy,
$\mathrm{CE}_{\text{base}}-\mathrm{CE}_{\text{patched}}$ (details in Appendix~\ref{app:ce-protocol}).
As shown in Figure~\ref{fig:ev-ce} (right), even removing a large number of unstable features produces only a modest loss
change, while masking far fewer stable features yields substantially larger degradation. 

Overall, unstable features are biased toward low-frequency surface-form patterns and have limited functional impact, whereas stable features capture most of the reconstruction- and prediction-relevant structure.

\subsection{Constructing SAE from Unique Features}
\label{sec:sae_sampling}

We next test whether stable cross-seed features can initialize seed-robust SAEs. We pool near-deduplicated decoder features from independently seeded SAEs, estimate each pooled feature's reappearance probability, and construct new $F=16{,}384$-latent SAEs from the most probable, least probable, or uniformly sampled pooled features before brief tuning; implementation details and probability histograms are in \Cref{app:ev-vs-mp,app:prob_hist}.

\begin{figure}[t]
\vspace{-1mm}
\centering
\includegraphics[width=0.85\linewidth]{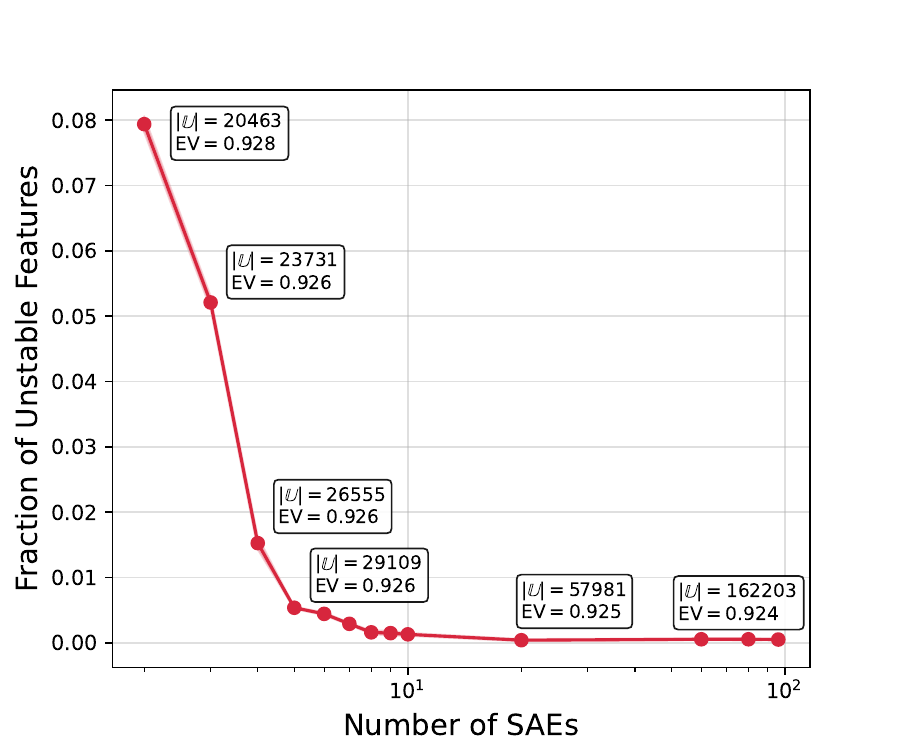}
\vspace{-2mm}
\caption{\textbf{Unstable-feature fraction in the most-probable construction as a function of the number of source SAEs.}
As the source pool grows, the selected dictionary contains progressively fewer unstable features and is already
dominated by stable ones for modest ensemble sizes. Boxes next to selected points report the size of the pooled
unique-feature set and the explained variance of the constructed SAE for that source-pool size.}
\label{fig:tuning}
\vspace{-2mm}
\end{figure}

Figure~\ref{fig:tuning} shows that only a modest number of source SAEs is needed: the most-probable construction quickly becomes dominated by high-probability features as the pool grows. After tuning, these dictionaries recover near-baseline explained variance, while least-probable dictionaries remain much worse and uniform sampling lies in between (Appendix~\ref{app:ev-vs-mp}). Thus stable pooled features can produce seed-robust SAEs without an apparent EV penalty. We further verify (Appendix~\ref{app:sae_bench}) that the most-probable construction SAE remains competitive with a standard TopK SAE on SAEBench metrics (Table~\ref{tab:saebench-stable-pooled}), indicating that the pooled construction improves feature stability without sacrificing downstream interpretability performance.

Appendix~\ref{app:stability-drift} shows that post-training a high-probability feature pool shifts some features toward lower reappearance probabilities, suggesting that lower-stability directions can be reconstruction-useful rather than mere random artifacts.

\section{Geometric Analysis}

We now investigate features geometrically at the level of decoder vectors
\(\boldsymbol{e}^{(s)}_j \in \mathbb{R}^d\). Our goal is to show that:
(i) unstable features concentrate in a lower effective-dimensional subspace than stable features, and
(ii) unstable features are linearly separable from the remaining features in decoder space, while the \emph{subspace} they span exhibits a high degree of cross-seed reproducibility.

\subsection{Effective Rank of Features Subspaces}

We compare the dimensionality of the decoder vector subspaces associated with stable and unstable
features. Given an index set \(\mathcal{I}^{(s)}_{\varepsilon}\) (either
\(\mathcal{U}^{(s)}_{\varepsilon}\) for unstable features or \(\mathcal{S}^{(s)}_{\varepsilon}\)
for stable features), define the submatrix of the decoder
\[
X^{(s)}_{\mathcal{I}_\varepsilon}
\;:=\;
\boldsymbol{W}^{(s)}_{\text{dec}}[:, \mathcal{I}^{(s)}_{\varepsilon}]
\in
\mathbb{R}^{d\times m_s},
\quad
m_s:=|\mathcal{I}^{(s)}_{\varepsilon}|.
\]

Using effective rank (ER; Appendix~\ref{sec:svd}) computed on the decoder submatrices
\(X^{(s)}_{\mathcal{U}_\varepsilon}\) and \(X^{(s)}_{\mathcal{S}_\varepsilon}\), unstable feature sets are consistently lower-dimensional: over $N=96$ seeds, $\mathrm{ER}/d \approx 0.59$--$0.65$ for unstable sets versus $\approx0.80$--$0.81$ for stable sets, a $20$--$27\%$ reduction (Table~\ref{tab:effective-rank-prob-sets}). Notably, such low-rank regions may reflect known sources of low-dimensional structure in transformers, including self-attention anisotropy and dimensional collapse in attention outputs \citep{godey2024anisotropyinherentselfattentiontransformers, wang2026dimensionalcollapsetransformerattention}.

\subsection{Overlap of Subspaces}

In this section we show that, although individual unstable features rarely reappear across seeds, the low-rank subspaces spanned by them are reproducible. This suggests that feature instability is primarily driven by seed-dependent mixing within a shared subspace, rather than changes in the subspace itself.

First, we train a logistic regression classifier to predict the binary label
\(y^{(s)}_i=\mathbf{1}\{i\in\mathcal{U}^{(s)}_{\varepsilon}\}\) from the corresponding decoder vector
\(\boldsymbol{e}^{(s)}_i\in\mathbb{R}^d\). In addition to within-seed evaluation, we test transferability by applying
a classifier trained on one seed to the decoder vectors of another seed.

We evaluate performance using the F1 score (Figure \ref{fig:f1_per_eps_neighbor}) as a function of \(N\), the number of
independently trained SAEs used to estimate the probabilities \(\hat{p}(\boldsymbol{f}^{(s)}_j)\). We find that the within-seed and
cross-seed F1 scores are very close, and for sufficiently large \(N\) they saturate
at approximately \(0.73\) for \(\varepsilon=0\) and  \(0.67\) for \(\varepsilon=0.1\). 

Next, to compare decoder vectors across seeds in more detail, we consider the explained variance $\mathrm{EV}^{(s)}_{\text{SVD}}(r)$ of singular values of matrix $X^{(s)}_{\mathcal{I}_\varepsilon}$ (formal definition in Appendix~\ref{sec:svd}),
which measures how well the top-\(r\) singular subspace captures the variance within the features set $\mathcal{I}_\varepsilon$.

Using the SVD of \(X^{(s)}_{\mathcal{I}_\varepsilon}\) and the subspace interpretation of \(\mathrm{EV}^{(s)}_{\text{SVD}}(r)\) (for details, see Appendix~\ref{sec:svd}), we define the cross-seed analogue \(\mathrm{EV}^{(a \to s)}_{\text{SVD}}(r)\) for \(s\neq a\), which measures how well the top-\(r\) singular subspace from seed \(a\) explains the feature subspace in seed \(s\).

\begin{figure}[!t]
\vspace{-1mm}
\centering
\includegraphics[width=0.85\linewidth]{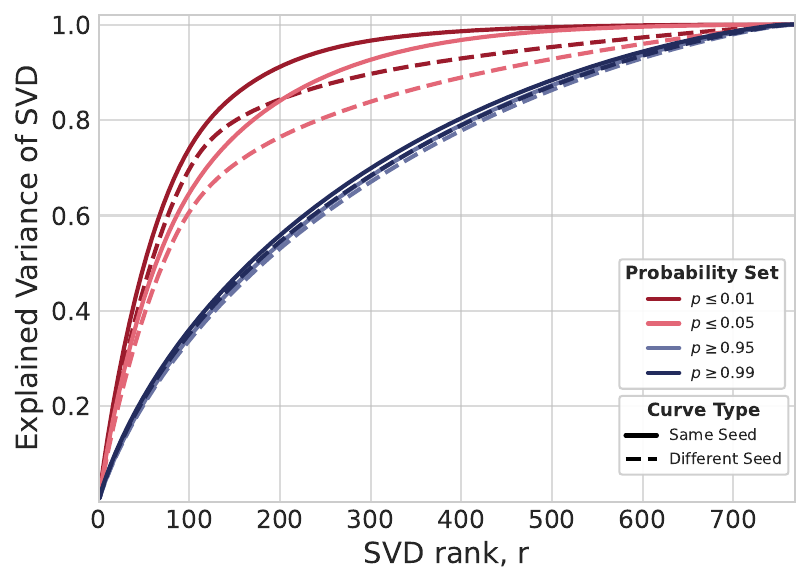}
\vspace{-2mm}
\caption{
\textbf{Explained variance of singular values of decoder submatrices versus SVD rank: within-seed (solid) and cross-seed transfer (dashed)}. This indicates that the subspaces learned in one seed accurately approximate those in other seeds.
}
\label{fig:unstable_overlap}
\vspace{-2mm}
\end{figure}

Figure \ref{fig:unstable_overlap} compares \(\mathrm{EV}^{(s)}_{\text{SVD}}(r)\) to \(\mathrm{EV}^{(a\to s)}_{\text{SVD}}(r)\), averaged over all choices of
$a$ and $s$ among $N=96$ seeds, for $\mathcal{U}_{\varepsilon}$ and $\mathcal{S}_{\varepsilon}$ with $\varepsilon \in\{0.01, 0.05\}$. We find that the top-$r$ singular subspace learned in one seed provides an accurate approximation to the corresponding feature subspaces in other seeds, including for unstable features.

\subsection{A Controlled Low-Rank Synthetic Model}
\label{sec:toy-model}

The preceding results suggest that unstable features may reflect basis ambiguity inside a shared low-dimensional region of
decoder space. To test whether this mechanism is sufficient, we construct a synthetic dictionary
\[
W
=
\begin{bmatrix}
D\\
UV
\end{bmatrix},
\qquad
D\in\mathbb{R}^{n_{\mathrm{full}}\times d},
\qquad
U\in\mathbb{R}^{n_{\mathrm{low}}\times r},
\]
\[
V\in\mathbb{R}^{r\times d},
\qquad
r<d,
\]
Rows of \(D\) are generic full-rank features, while rows of \(UV\) lie in the shared subspace
\(\mathrm{rowspan}(V)\). We generate activations by sampling \(k\) dictionary rows, summing them with unit coefficients,
and training TopK SAEs from multiple seeds. Full-rank features should be individually identifiable, whereas low-rank
features may only be identifiable up to rotations or mixtures within the shared span.

\begin{figure}[t]
\vspace{-1mm}
\centering
\includegraphics[width=\linewidth]{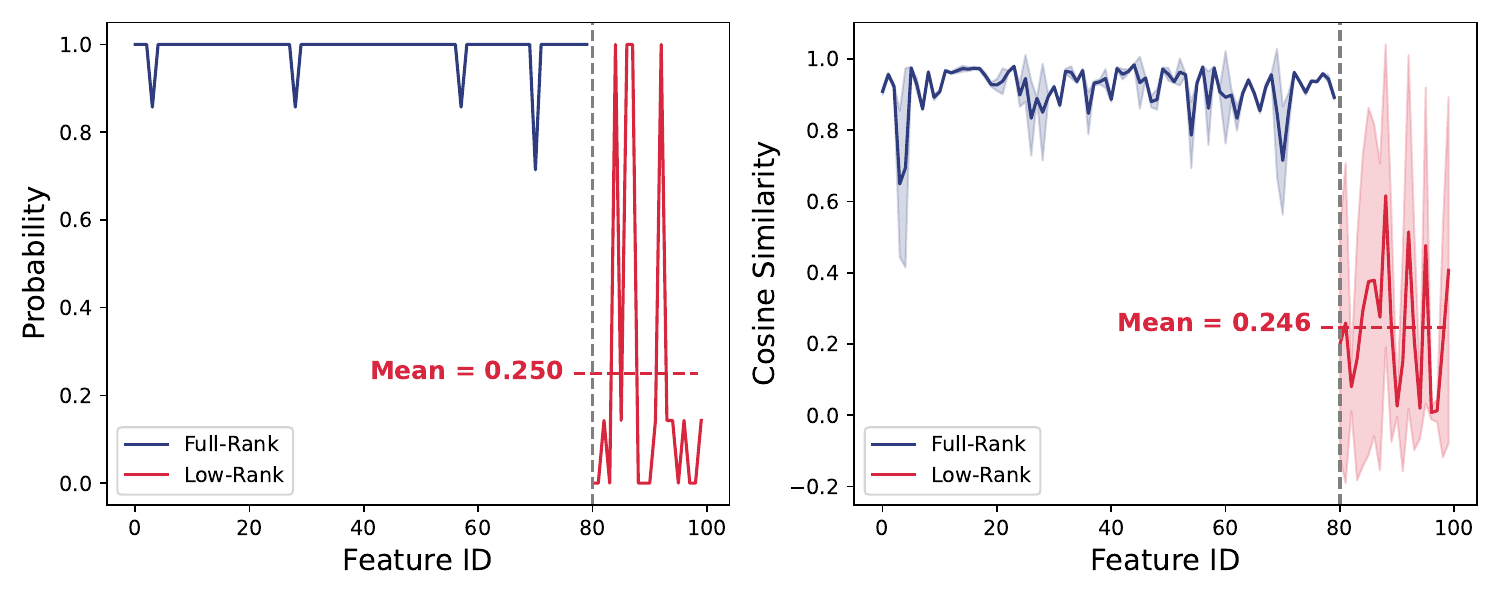}
\vspace{-2mm}
\caption{
\textbf{Synthetic low-rank model ($d=32$, $r=2$, $k=8$).}
The first $80$ ground-truth features are full-rank features and the last $20$ lie in a shared rank-$2$ subspace.
Full-rank features (blue) have near-perfect cross-seed reappearance probability and cosine similarity to their matched
ground-truth features, whereas low-rank features (red) do not.
}
\label{fig:toy_prob_cossim}
\vspace{-2mm}
\end{figure}

Figure~\ref{fig:toy_prob_cossim} confirms this prediction for \(d=32,r=2,k=8\): full-rank features have near-perfect
reappearance probability and cosine similarity to matched ground-truth features, while low-rank features have much lower
reappearance probability (\(\approx0.25\)) and worse one-to-one recovery. Appendix~\ref{app:toy-model} shows the same
qualitative split in additional settings and further shows that the learned low-rank block has lower effective rank and
non-random cross-seed subspace similarity. Thus, the empirical geometry of unstable SAE features can arise naturally from
seed-dependent basis choices within a reproducible low-rank subspace, rather than from the absence of shared structure.
Appendix~\ref{app:manual-identification} gives an additional residual-based diagnostic that partially recovers the
low-rank ground-truth atoms by clustering examples with one active unstable feature.


\section{Other Setups}
\label{sec:other-setups}

We next test whether the stability patterns from the main setting persist under changes to the base model, layer, and
dictionary size, and under alternative SAE nonlinearities, training budgets, and a random-model null baseline.

Overall, these ablations show that a non-trivial unstable subset coexists with a large stable subset across settings.

\textbf{Across base models, layers, and dictionary sizes.}
We first train TopK SAEs across several base models, layers, and dictionary sizes.
Across all evaluated settings, we observe a substantial stable subset alongside a non-trivial unstable subset, indicating
that seed dependence is a general property of this training objective rather than a one-off artifact.
Stability varies across settings: for GPT-2 and Gemma-2, instability generally decreases with depth, while some
Pythia settings deviate from this trend.
Full endpoint fractions for all configurations are reported in Appendix~\ref{app:other-setups} (Table~\ref{tab:endpoints-topk} and Figure~\ref{fig:evolution_of_stability}). Figure~\ref{fig:evolution_of_stability} shows the full GPT-2 layer-wise trajectory, where stable features increase with depth, unstable features decrease, and next-layer matches occur predominantly among stable features.

\textbf{Effect of sparsity mechanism and reconstruction quality.}
\Cref{tab:gpt2-sae-type-layer7} compares SAE variants on GPT-2 layer 7 at fixed $F=2^{14}$, including EV.
TopK and BatchTopK are nearly indistinguishable both in stability and EV ($\approx 0.892$), suggesting that the dominant
source of seed dependence is the hard $k$-sparsity constraint itself rather than whether it is imposed per-token or per-batch. Within TopK SAEs, increasing \(k\) improves EV but monotonically increases the unstable fraction and decreases the stable fraction, consistent with stronger sparsity constraints reducing seed-dependent degrees of freedom.
Vanilla (ReLU+$\ell_1$) is extremely stable (near-zero unstable fraction), but at comparable sparsity it has substantially
lower EV than TopK, highlighting a clear stability--reconstruction tradeoff when moving along the choice of sparsity/activation mechanism.
HierarchicalTopK also improves stability relative to TopK but lowers EV, consistent with additional structure in the sparsity objective reducing seed dependence at some reconstruction cost.
JumpReLU improves stability relative to TopK while retaining similar EV at higher sparsities, suggesting that different sparsity mechanisms can shift this stability--reconstruction balance in different ways.

\begin{figure}[t]
\vspace{-1mm}
\centering
\includegraphics[width=0.85\linewidth]{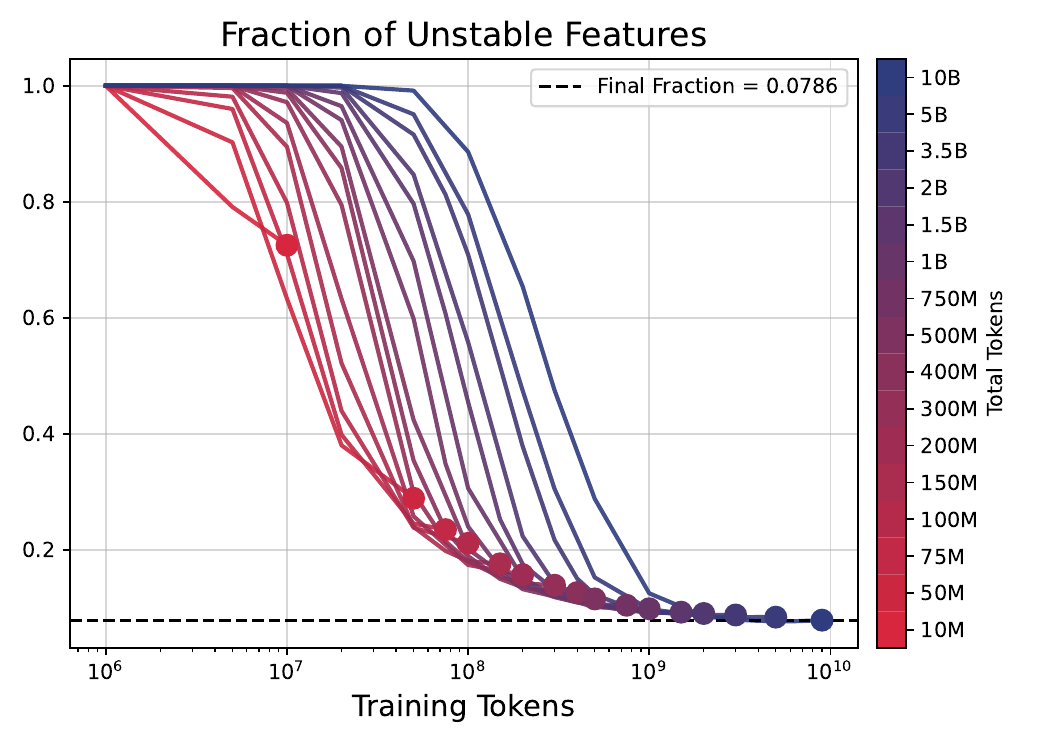}
\vspace{-2mm}
\caption{
\textbf{Unstable fraction vs.\ SAE training tokens.}
Fraction of unstable features in the main TopK setting as a function of total SAE training tokens.
}
\label{fig:unstable_vs_tokens}
\vspace{-2mm}
\end{figure}

\textbf{More SAE training does not remove instability.}
\Cref{fig:unstable_vs_tokens} shows the unstable fraction in the main TopK setting as a function of total SAE training
tokens from $10$M to $10$B.
The unstable fraction decreases early in training and then approaches a clear non-zero plateau ($\approx 8\%$ at $10$B
tokens in this setting).
In contrast, the stable fraction continues to increase with additional training but with diminishing returns; we report the
corresponding curve and full setup details in Appendix (Fig.~\ref{fig:stable_vs_tokens}).

\begin{figure}[t]
\vspace{-1mm}
\centering
\includegraphics[width=0.8\linewidth]{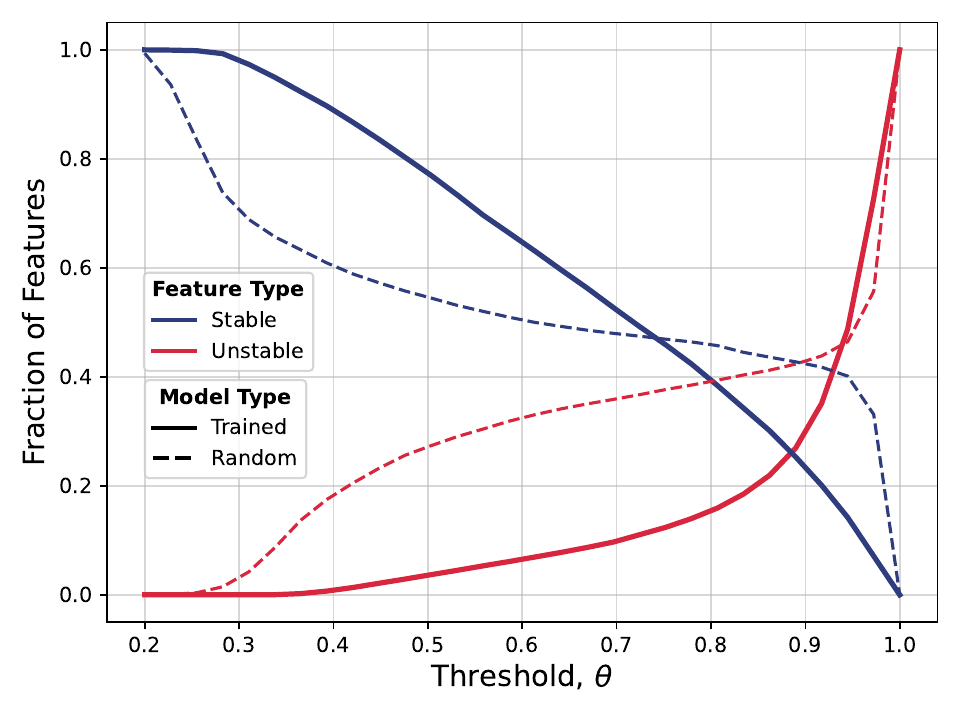}
\vspace{-2mm}
\caption{
\textbf{Dead-salmon control: stability on trained vs.\ random transformers.}
Fractions of stable and unstable features as a function of cosine matching threshold $\theta$,
comparing SAEs trained on a trained GPT-2 versus a randomly initialized GPT-2 (same architecture and SAE setup).
}
\label{fig:theta-random-trained}
\vspace{-2mm}
\end{figure}

\textbf{Dead-salmon baseline: trained vs.\ random transformers.}
Finally, we repeat the same SAE training procedure on activations from a randomly initialized GPT-2 (same architecture and
SAE hyperparameters).
Our stability metric sharply distinguishes these cases: \Cref{fig:theta-random-trained} compares stability as a function of
the cosine matching threshold $\theta$ for SAEs trained on a trained GPT-2 versus a random GPT-2.
Across a wide range of $\theta$, the trained-model SAEs retain a large stable subset, whereas in the random-model setting
the stable fraction collapses and instability dominates.
Importantly, automatic interpretation alone would not reliably flag this failure mode: SAEs trained on the random model
can still achieve high detection scores (Appendix~\ref{app:autointerp-random}, Fig.~\ref{fig:autointerp_random}).
Together, these results show why stability is useful alongside automatic interpretation: plausible explanations can arise even when features are not reproducible across seeds.

\begin{table}[t]
\centering
\small
\setlength{\tabcolsep}{5pt}
\begin{tabular}{llccc}
\toprule
SAE type & Sparsity & EV & Unstable & Stable \\
\midrule
\multirow{2}{*}{Vanilla (ReLU + $\ell_1$)}
& 62 & 0.833 & 0.003 & 0.810 \\
& 91 & 0.856 & 0.006 & 0.764 \\
\midrule
\multirow{4}{*}{TopK}
& 32 & 0.862 & 0.053 & 0.619 \\
& 48 & 0.882 & 0.080 & 0.560 \\
& 64 & 0.892 & 0.098 & 0.522 \\
& 80 & 0.902 & 0.118 & 0.499 \\
\midrule
BatchTopK
& 64 & 0.892 & 0.097 & 0.523 \\
\midrule
HierarchicalTopK
& 64 & 0.876 & 0.028 & 0.716 \\
\midrule
\multirow{3}{*}{JumpReLU}
& 29 & 0.841 & 0.049 & 0.624 \\
& 58 & 0.884 & 0.034 & 0.620 \\
& 93 & 0.902 & 0.038 & 0.602 \\
\bottomrule
\end{tabular}
\caption{SAE type comparison on GPT-2 at layer 7 with $F=2^{14}$.}
\label{tab:gpt2-sae-type-layer7}
\end{table}

\section{Conclusion}

Our results show that SAE seed dependence is structured rather than arbitrary. Feature stability, measured via cross-run reappearance probability, separates learned features into two empirically distinct regimes: stable features carry most of the reconstruction- and prediction-relevant signal and more often correspond to structural or compositional patterns, while unstable features have weaker marginal impact and are biased toward low-frequency surface-form triggers.

Crucially, unstable features are not merely failed or noisy latents. Although individual unstable features rarely reappear across seeds, they concentrate in reproducible lower-rank subspaces, suggesting that seed dependence often reflects different basis choices within a shared region of activation space. Our controlled toy model makes this mechanism explicit: low-rank ground-truth features can be recovered at the subspace level while remaining non-identifiable as individual SAE latents across seeds.

Taken together, these results show that a single SAE dictionary can obscure important cross-seed structure. They also show that cross-seed structure can be reused: by pooling high-probability features across runs and briefly post-training, we construct more seed-robust dictionaries without sacrificing reconstruction quality. A natural direction for future work is to identify the model components that give rise to these reproducible low-rank components, and to develop methods for recovering identifiable individual features within them.

\section*{Limitations}

First, the stable and unstable sets depend on two threshold choices: the decoder-cosine matching threshold $\theta$ for cross-seed recurrence and the endpoint cutoff $\varepsilon$ for selecting features that reappear almost never or almost always. In the main experiments we use $\theta=0.7$ and $\varepsilon=0.05$. Our threshold analyses indicate that the main qualitative trends are robust to reasonable variations of both, but the precise sizes and membership of the endpoint sets remain conditional on these choices. Second, our low-rank subspace analysis should be interpreted as evidence for a concrete mechanism of seed dependence, rather than as a complete identification of the mechanism generating instability in real LLM SAEs. The empirical results and synthetic model show that individually unstable features can arise from basis ambiguity within reproducible lower-rank subspaces, but they do not prove that every unstable feature in real models is produced by this mechanism. Third, our no stability–EV tradeoff result is specific to the feature-pool construction, which uses features aggregated from several independently trained SAEs. Thus, it shows that the trade-off can be avoided in this construction setting, but not that a single SAE training objective can achieve the same stability profile at no reconstruction cost.

\bibliography{custom}

\newpage
\appendix

\section{SAE Training Details} \label{sec:appendix:training_details}

\subsection{SAE variants}

\emph{Vanilla SAE} \citep{cunningham2023sparse} uses $\mathrm{ReLU}$ as the nonlinearity and $\mathcal{L}_{\text{reg}}=\|\boldsymbol{z}\|_1$.
\emph{TopK SAE} \citep{gao2025scaling} imposes sparsity on $\boldsymbol{z}$ by zeroing out all entries outside the top-$k$ values.
\emph{Batch TopK SAE} \citep{bussmann2024batchtopksparseautoencoders} applies the same idea at the batch level, zeroing out elements below the top-(batch size $\times$ k) threshold; these activations fix the desired sparsity level and remove the need for an explicit regularization term.
\emph{HierarchicalTopK SAE} \citep{balagansky-etal-2025-train} trains a single dictionary across nested sparsity budgets, so increasingly large prefixes of the selected features are optimized to reconstruct the activation.
\emph{JumpReLU SAE} \citep{rajamanoharan2024jumpingahead} modifies the Vanilla SAE by learning a threshold $\theta_i\in\mathbb{R}$ for each feature and using $\mathrm{JumpReLU}(\boldsymbol{x})=\boldsymbol{x}\mathrm{H}(\boldsymbol{x}-\theta)$, where $\mathrm{H}$ is the Heaviside function.

\subsection{SAE training setup}
\label{app:sae-training-setup}

Unless otherwise stated, each SAE is trained on \(1\)B token activations sampled from the \texttt{sample-10BT} split of FineWeb. We train all seeds with the same activation sampling order: at each optimization step, we maintain an activation store containing \(8192\times 32\) token activations and sample \(8192\) token positions uniformly at random from this store. The random seed controlling this sampling procedure is shared across SAE runs, so differences between runs are due to SAE initialization only.

We use an auxiliary loss for dead features, following \citet{gao2025scaling}. A feature is marked as \emph{dead} if it has not activated for more than \(20\) consecutive batches. The auxiliary loss gives such dead features a learning signal by asking them to help reconstruct the residual error left by the main active features. Concretely, after computing the standard SAE reconstruction \(\hat{\boldsymbol{x}}\), we form the residual
\[
\boldsymbol{r}=\boldsymbol{x}-\hat{\boldsymbol{x}}.
\]
The auxiliary loss selects dead features with large activations on the current batch and uses their decoder directions to reconstruct this residual. This encourages inactive features to specialize to currently unexplained directions, reducing the chance that they remain permanently unused. The auxiliary term is then added to the standard SAE training objective during training.

\subsection{Decoder normalization}

During training, we normalize decoder weights so that feature embeddings and gradients have unit norm. Namely, suppose that $w_i$ and $g_i$ is a decoder column and its gradient correspondingly, then the procedure is:

\begin{enumerate}
    \item Compute row-normalized decoder weights: $u_i = \frac{w_i}{\|w_i\|_2}$.
    \item Remove the component parallel to $u_i$ from the gradient: $g_i \leftarrow g_i - (g_i^Tu_i)u_i$, so that gradient step will be performed on the tangent space of the unit sphere.
    \item Set weights to normalized: $w_i \leftarrow u_i$.
\end{enumerate}

After training, we fold the mean and the standard deviation of the initial hidden state buffer into the weights so that no explicit normalization of the input and rescaling of the output is required, then normalize the decoder and scale encoder weights and bias to counteract this normalization, so that SAE do not change the output.


\section{Additional Details for Quantitative Analysis}
\label{app:quant-details}

\subsection{Evaluation tokens and activation collection}
\label{app:quant-eval}

All quantitative statistics (frequencies, magnitudes, entropy, EV and CE under patching) are computed over a held-out
collection of token positions.
Let $B_{\mathrm{eval}}$ be the number of evaluation batches and $T$ the context length, and define
\[
N_{\mathrm{tok}} \;=\; B_{\mathrm{eval}}\cdot T.
\]
For token index $n\in\{1,\dots,N_{\mathrm{tok}}\}$, let $\boldsymbol{h}_n\in\mathbb{R}^d$ denote the base-model activation
at the SAE training location (residual stream), and let $\boldsymbol{z}_n\in\mathbb{R}^F_{\ge 0}$ denote SAE feature
activations.
We use the SAE reconstruction
\begin{equation}
\hat{\boldsymbol{h}}_n
=
\boldsymbol{W}_{\mathrm{dec}}\boldsymbol{z}_n+\boldsymbol{b}_{\mathrm{dec}}.
\label{eq:sae-recon-quant-app}
\end{equation}
All evaluation batches are disjoint from SAE training batches.

\subsection{Cosine-threshold sensitivity across SAE families}
\label{app:threshold-sensitivity}

\begin{figure}[t]
\vspace{-1mm}
\centering
\includegraphics[width=\linewidth]{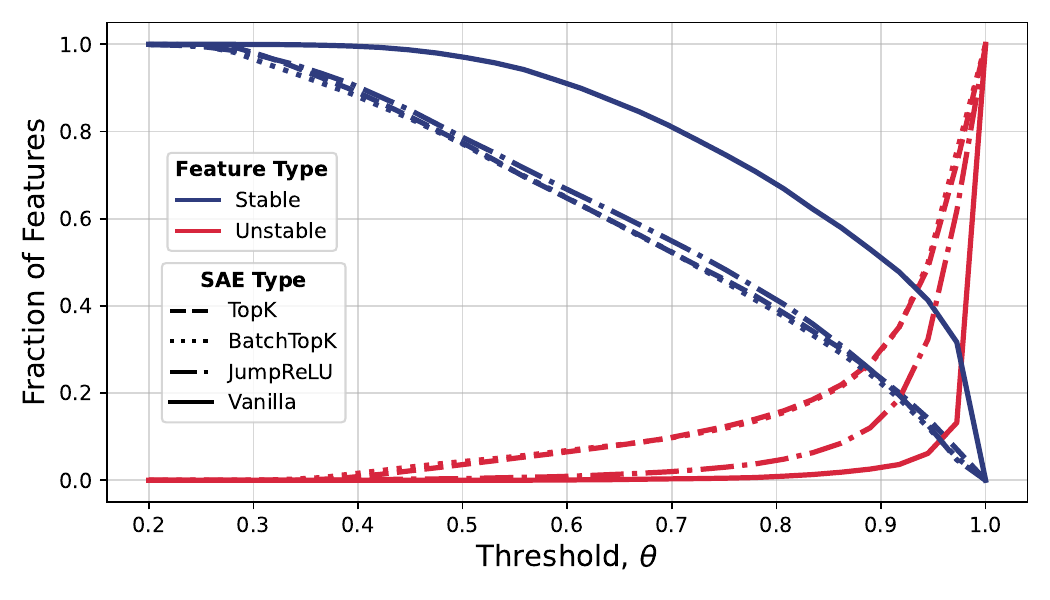}
\vspace{-2mm}
\caption{\textbf{Endpoint fractions vs.\ cosine threshold $\theta$.} Fractions of \emph{stable} and \emph{unstable}
features as a function of cosine matching threshold $\theta$ for several SAE types in the same base-model setting.}
\label{fig:theta-sweep}
\vspace{-2mm}
\end{figure}

We vary the cosine matching threshold $\theta$ used to declare cross-seed feature matches and recompute the endpoint
fractions for each SAE family in the main GPT-2 setting.
Figure~\ref{fig:theta-sweep} shows that increasing $\theta$ makes matching stricter, so the stable fraction falls and the
unstable fraction rises across all SAE types.
The qualitative ordering is preserved throughout the sweep: TopK and BatchTopK remain nearly identical, JumpReLU stays
intermediate, and Vanilla exhibits the strongest endpoint stability.

\subsection{Usage statistics: activation frequency and mean magnitude}
\label{app:usage-stats}

\begin{figure}[t]
\vspace{-1mm}
\centering
\includegraphics[width=\linewidth]{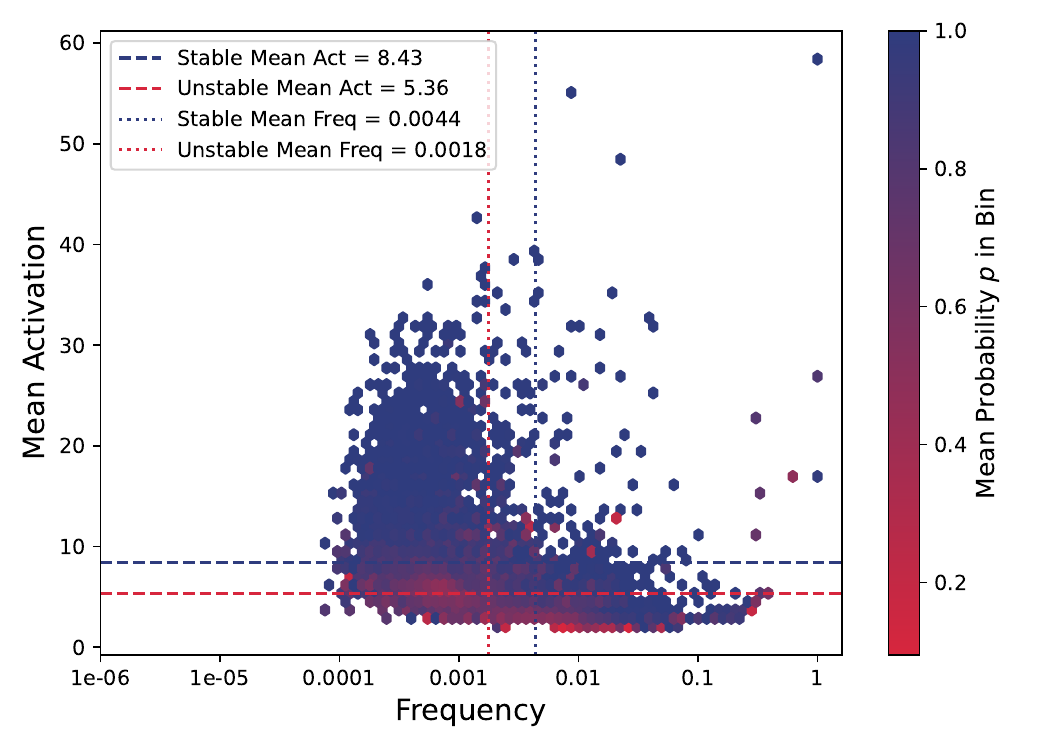}
\vspace{-2mm}
\caption{\textbf{Activation frequency and conditional mean magnitude for stable vs.\ unstable features.} Unstable
features are concentrated at lower frequencies, while stable features exhibit a heavier high-magnitude tail.}
\label{fig:mean-act-freq}
\vspace{-2mm}
\end{figure}

We quantify how often each feature is used and how strongly it fires when active via:
\begin{align}
\omega_i
&=
\frac{1}{N_{\mathrm{tok}}}\sum_{n=1}^{N_{\mathrm{tok}}}\mathbf{1}\{z_{n,i}>0\},
\label{eq:omega}
\\
\mu_i
&=
\frac{\sum_{n=1}^{N_{\mathrm{tok}}} z_{n,i}}
     {\sum_{n=1}^{N_{\mathrm{tok}}}\mathbf{1}\{z_{n,i}>0\}}.
\label{eq:mu}
\end{align}
Here $\omega_i$ is the fraction of evaluated token positions where feature $i$ activates, and $\mu_i$ is the mean
activation magnitude conditioned on activation. Figure~\ref{fig:mean-act-freq} visualizes these two statistics jointly
for stable and unstable features.

\subsection{Token entropy computation}
\label{app:token-entropy}

Let $v_n\in\{1,\dots,|\mathcal{V}|\}$ be the vocabulary token id at token position $n$.
For each feature $i$, define token-conditional activation counts and the induced token distribution
\begin{align}
c_i(v)
&=
\sum_{n=1}^{N_{\mathrm{tok}}}\mathbf{1}\{(v_n=v)\wedge(z_{n,i}>0)\},
\label{eq:token-counts}
\\
\pi_i(v)
&=
\frac{c_i(v)}{\sum_{v'\in\mathcal{V}} c_i(v')},
\label{eq:token-dist}
\end{align}
and token entropy
\begin{equation}
H_i
=
-\sum_{v\in\mathcal{V}} \pi_i(v)\log \pi_i(v).
\label{eq:token-entropy}
\end{equation}
We use the natural logarithm.
Features that never activate on the evaluation set are excluded from entropy plots.

\subsection{Masking protocol, uncertainty, and reweighting}
\label{app:masking-protocol}

For a masked index set $M\subseteq\{1,\dots,F\}$, define
\begin{align}
\boldsymbol{z}^{(-M)}_n
&=
\boldsymbol{z}_n\odot(\boldsymbol{1}-\boldsymbol{1}_M),
\label{eq:masked-code}
\\
\hat{\boldsymbol{h}}^{(-M)}_n
&=
\boldsymbol{W}_{\mathrm{dec}}\boldsymbol{z}^{(-M)}_n+\boldsymbol{b}_{\mathrm{dec}}.
\label{eq:masked-recon}
\end{align}

We measure reconstruction quality by explained variance
\begin{equation}
\mathrm{EV}(M)
=
1
-
\frac{\mathbb{E}_n\!\left[\|\boldsymbol{h}_n-\hat{\boldsymbol{h}}^{(-M)}_n\|_2^2\right]}
     {\mathbb{E}_n\!\left[\|\boldsymbol{h}_n-\bar{\boldsymbol{h}}\|_2^2\right]},
\,
\bar{\boldsymbol{h}}=\mathbb{E}_n[\boldsymbol{h}_n].
\label{eq:ev}
\end{equation}

For each masking budget $N$, we sample masks as follows:
\begin{itemize}
\item Sample $M_S$ by drawing $N$ features uniformly without replacement from $\mathcal{S}_\varepsilon$.
\item Sample $M_U$ by drawing $4N$ features uniformly without replacement from $\mathcal{U}_\varepsilon$.
\end{itemize}
We repeat this sampling procedure $R=10$ times independently for each $N$.
Plots report the mean across these $R$ draws, and the shaded band indicates one standard deviation across draws.

\textbf{Reweighting.}
In addition to direct masked reconstructions $\hat{\boldsymbol{h}}^{(-M)}_n$, we evaluate a reweighted variant that matches
the $\ell_2$ norm of the unmasked SAE reconstruction $\hat{\boldsymbol{h}}_n$ at each token position:
\begin{equation}
\tilde{\boldsymbol{h}}^{(-M)}_n
=
\hat{\boldsymbol{h}}^{(-M)}_n\cdot
\frac{\|\hat{\boldsymbol{h}}_n\|_2}{\|\hat{\boldsymbol{h}}^{(-M)}_n\|_2+\delta},
\label{eq:reweight}
\end{equation}
where $\delta>0$ is a small constant for numerical stability.

\subsection{Next-token loss under activation patching}
\label{app:ce-protocol}

To measure downstream impact, we patch the residual stream at the SAE training location.
For each token position $n$, we run a clean forward pass and compute cross-entropy $\mathrm{CE}_{\text{base}}$ on the
evaluation tokens.
We then rerun the forward pass where $\boldsymbol{h}_n$ is replaced by either
$\hat{\boldsymbol{h}}^{(-M)}_n$ (no reweighting) or $\tilde{\boldsymbol{h}}^{(-M)}_n$ (with reweighting), and compute
$\mathrm{CE}_{\text{patched}}(M)$ on the same evaluation tokens.
We report
\begin{equation}
\Delta \mathrm{CE}(M)
=
\mathrm{CE}_{\text{base}}
-
\mathrm{CE}_{\text{patched}}(M),
\label{eq:delta-ce}
\end{equation}
so more negative values indicate larger degradation (as in Figure~\ref{fig:ev-ce}).


\section{Additional Details for Constructed and Stable-Pooled SAEs}
\label{app:constructed-saes}

\subsection{Explained variance during tuning of constructed SAEs}
\label{app:ev-vs-mp}

To test feature-pool construction, we train a collection of independently seeded SAEs and merge their decoder features into the unique pool $\mathbb{U}$, so near-duplicate cross-seed features appear only once.
Formally, we process all features from \(S\) source SAEs in lexicographic order.
Let \(t=(s-1)F+i\) index candidate \(c_t=\boldsymbol{f}^{(s)}_i\), and write \(\boldsymbol{e}(c_t)=\boldsymbol{e}^{(s)}_i\).
Starting with \(\mathbb{U}^{(0)}=\varnothing\), define
\begin{equation}
\label{eq:unique_features}
\mathbb{U}^{(t)}
=
\begin{cases}
\mathbb{U}^{(t-1)}\cup\{c_t\},
&
\displaystyle
\max_{\boldsymbol{f}\in\mathbb{U}^{(t-1)}}
\boldsymbol{e}(c_t)^\top \boldsymbol{e}(\boldsymbol{f}) < \theta,
\\[1.2ex]
\mathbb{U}^{(t-1)},
&
\text{otherwise.}
\end{cases}
\end{equation}
where the maximum over the empty set is \(-\infty\).
The final pool is \(\mathbb{U}=\mathbb{U}^{(SF)}\), and we use the same threshold \(\theta=0.7\) as in feature matching.
For a source-pool size $n$, we form the pool from the first $n$ source SAEs and estimate each pooled feature's empirical reappearance probability $\hat p$ within that subset.
We then instantiate new SAEs with $F=16{,}384$ latents using three feature-selection rules: the $F$ most probable pooled features, the $F$ least probable pooled features, and an \emph{equiprobable} baseline that samples pooled features uniformly regardless of $\hat p$.

\begin{figure}[t]
\vspace{-1mm}
\centering
\includegraphics[width=0.78\linewidth]{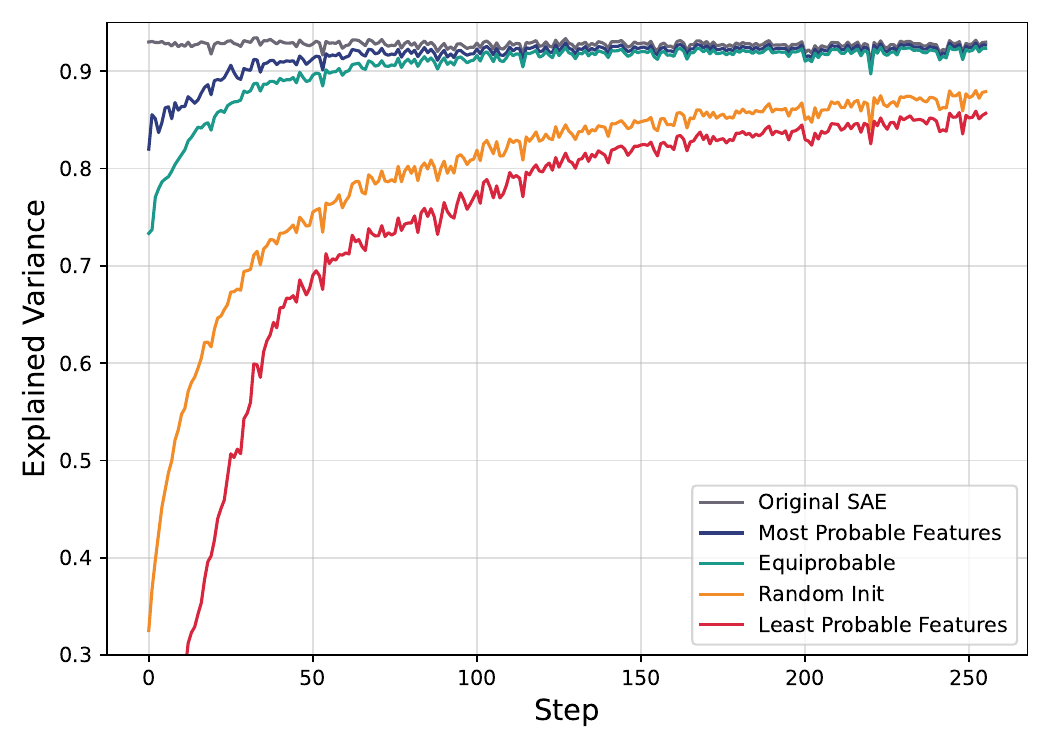}
\vspace{-2mm}
\caption{
\textbf{Explained variance during brief tuning of SAEs constructed from different feature subsets.}
Initialization from the most probable features recovers nearly the same explained variance as the original SAE, the
equiprobable construction performs somewhat worse, and the least-probable construction lags substantially behind.
}
\label{fig:ev_vs_mp}
\vspace{-2mm}
\end{figure}

All constructed SAEs are briefly tuned for approximately \(2\)M tokens before evaluation. Figure~\ref{fig:ev_vs_mp} reports the explained variance during this brief tuning for SAEs initialized from the most-probable, equiprobable, and least-probable feature subsets. The least-probable construction remains below even a randomly initialized SAE trained for the same token budget, whereas the most-probable and equiprobable constructions reach essentially the same EV as the post-trained baseline SAE.

\subsection{Probability distributions of selected features in constructed SAEs}
\label{app:prob_hist}

\begin{figure}[t]
    \centering
    \includegraphics[width=\linewidth]{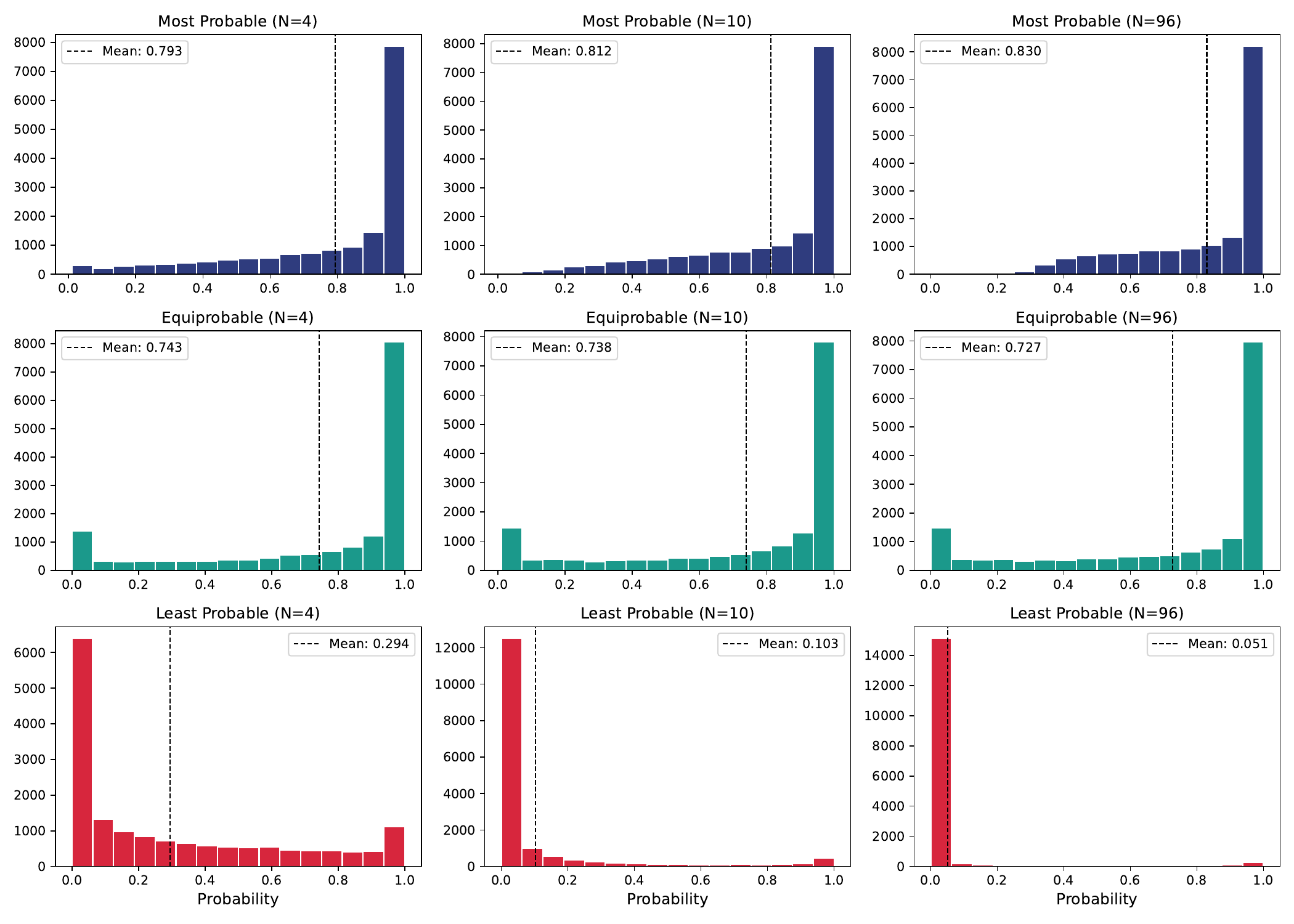}
    \caption{\textbf{Probability distributions of selected features for different source-pool sizes.}
    Rows correspond to the most-probable, equiprobable, and least-probable constructions; columns use $n=4$, $10$, and
    $96$ source SAEs. Even small source pools already make the most-probable construction concentrate on high-reappearance
    features, while the least-probable construction remains dominated by near-zero-probability features.}
    \label{fig:pool_prob_hist}
\end{figure}

Figure~\ref{fig:pool_prob_hist} shows the reappearance-probability distributions induced by the most-probable,
equiprobable, and least-probable construction rules across different source-pool sizes.

\subsection{SAEBench metrics for standard and most-probable construction SAEs}
\label{app:sae_bench}

Table~\ref{tab:saebench-stable-pooled} shows that the most-probable construction SAE remains broadly competitive with standard TopK SAEs on SAEBench: SCR and TPP decrease modestly, while Sparse Probing Top-1 and the Autointerp score slightly improve.

\begin{table}[t]
\centering
\small
\setlength{\tabcolsep}{6pt}
\begin{tabular}{lcc}
\toprule
Metric & Standard & Most-probable \\
\midrule
Sparse Probing (Top-1) & 0.656 & 0.673 \\
AutoInterp (mean) & 0.818 & 0.850 \\
SCR (Top 20) & 0.269 & 0.258 \\
TPP (Top 20) & 0.107 & 0.090 \\
\bottomrule
\end{tabular}
\caption{SAEBench metrics for standard and most-probable construction SAEs.}
\label{tab:saebench-stable-pooled}
\end{table}

\subsection{Automatic interpretation for baseline vs.\ most-probable-feature SAE}
\label{app:autointerp-sampling}

\begin{figure}[t]
\vspace{-1mm}
\centering
\includegraphics[width=\linewidth]{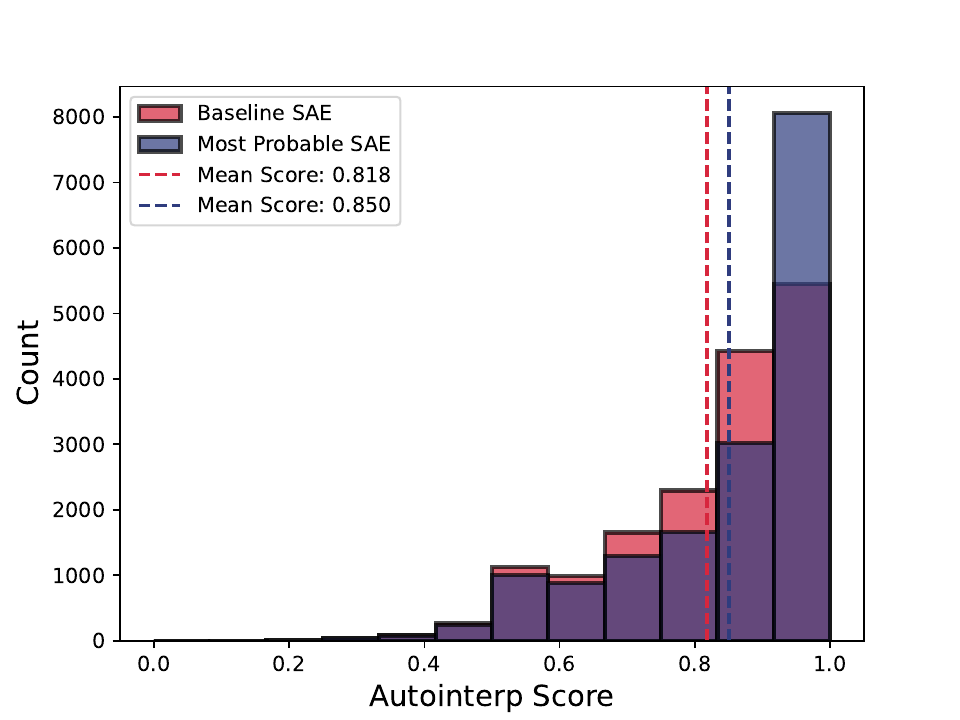}
\vspace{-2mm}
\caption{\textbf{Automatic interpretation scores: baseline SAE vs.\ SAE constructed from the most probable features.}
Histogram of detection scores; dashed lines show mean scores for each SAE.}
\label{fig:autointerp_sampling}
\vspace{-2mm}
\end{figure}

Figure~\ref{fig:autointerp_sampling} compares detection-score distributions for the baseline SAE and the SAE constructed
from the most probable feature pool.

\section{Additional Discussion on Stability-Inducing Methods}
\label{app:stability-methods}
\label{app:stability-drift}

Our main analysis studies what unstable features are like and why they arise under a fixed SAE objective. A related but distinct question is how to make learned SAE dictionaries more stable. We group stability-inducing approaches into three broad axes: explicit or implicit regularization, metric reweighting, and bagging-style feature pooling.

\textbf{Regularization and architectural constraints.}
Regularization or additional constraints can reduce the space of near-equivalent solutions and thereby increase cross-seed agreement. In our experiments, Vanilla ReLU+$\ell_1$ SAEs are substantially more stable than TopK SAEs at comparable sparsity, but have lower EV. JumpReLU SAEs also include an $\ell_1$-style sparsity regularization term, which may partly explain their improved stability relative to TopK. Archetypal SAEs constrain the optimization by tying learnable features to archetypes derived from training data \citep{fel2025archetypal}; we do not include them in our quantitative comparison because the original method was evaluated on vision models, and in our text-activation experiments we were unable to obtain competitive reconstruction performance. HierarchicalTopK is different: it does not add an explicit $\ell_1$ penalty, but can be viewed as an implicit regularizer because a single dictionary must support good reconstruction across nested sparsity budgets \citep{balagansky-etal-2025-train}. This may reduce the degeneracy of the fixed-\(k\) TopK objective and make seed-dependent basis choices less likely. This is also consistent with the TopK sparsity sweep in Table~\ref{tab:gpt2-sae-type-layer7}: lower \(k\) yields lower EV but higher stability, suggesting that stronger sparsity constraints can reduce seed-dependent basis choices.

\begin{figure}[t]
\centering
\includegraphics[width=\linewidth]{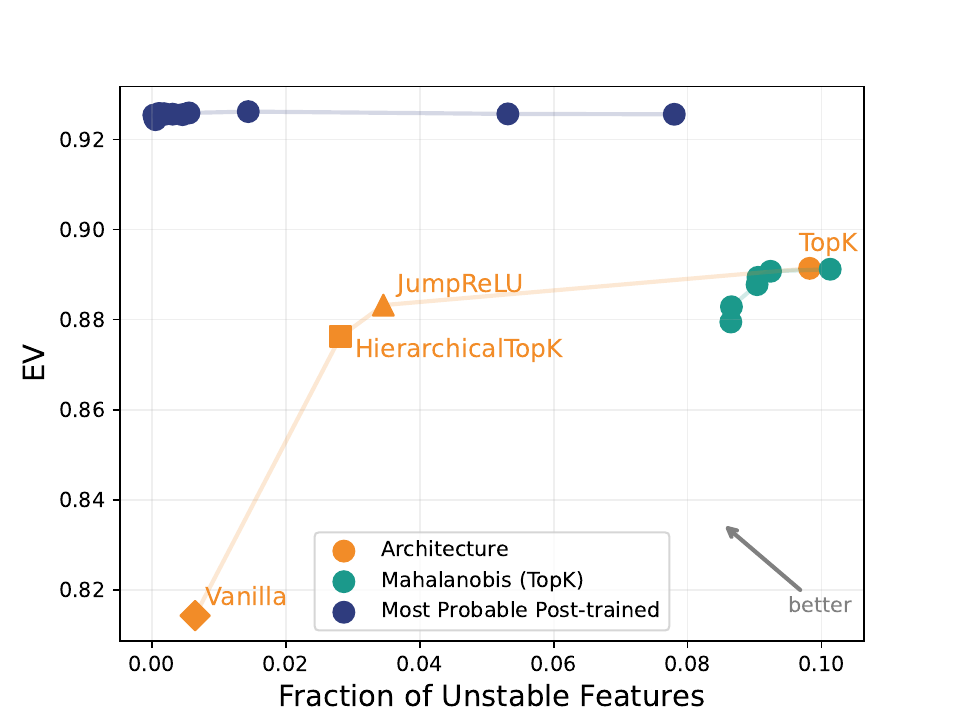}
\caption{
\textbf{Routes to SAE stability in the EV--instability plane.}
Orange points compare SAE architecture/objective variants, including TopK, JumpReLU, Vanilla, and HierarchicalTopK. Green points vary the Mahalanobis/whitening interpolation for TopK SAEs. Blue points show most-probable feature-pool SAEs after post-training with different numbers of source SAEs; rightmost blue point corresponds to the post-trained TopK SAE (i.e. feature pool with single SAE). Architectural and loss-level changes can reduce instability but trade off against EV, whereas feature pooling can reach low instability while preserving high EV.
}
\label{fig:ev_unstable_tradeoff}
\end{figure}

\textbf{Data whitening and Mahalanobis losses.}
Our results show that unstable features contribute less to reconstruction, suggesting that they may lie in directions that are downweighted by the standard MSE objective. This motivates reweighting the reconstruction metric. Let \(\boldsymbol{x}\in\mathbb{R}^d\) be a hidden state, \(\hat{\boldsymbol{x}}\) its SAE reconstruction, \(\boldsymbol{r}=\hat{\boldsymbol{x}}-\boldsymbol{x}\) the residual, and \(\Sigma\in\mathbb{R}^{d\times d}\) the activation covariance. Standard MSE uses \(\boldsymbol{r}^{\top}\boldsymbol{r}\), while a regularized Mahalanobis loss uses
\[
\boldsymbol{r}^{\top}(\Sigma+\lambda I)^{-1}\boldsymbol{r}.
\]
This is closely related to training on whitened activations \citep{saraswatula2025datawhiteningimprovessparse}. To interpolate continuously between MSE and this Mahalanobis objective, we train with
\[
\boldsymbol{r}^{\top}(\Sigma+\lambda I)^{-\alpha}\boldsymbol{r},
\qquad \alpha\in[0,1],
\]
where \(\alpha=0\) recovers MSE and \(\alpha=1\) recovers regularized Mahalanobis loss. Figure~\ref{fig:ev_unstable_tradeoff} shows that increasing \(\alpha\) can reduce instability, but substantially worsens EV. This is expected: reweighting low-variance directions makes the training objective less aligned with standard reconstruction quality.

\begin{figure}[t]
\centering
\includegraphics[width=\linewidth]{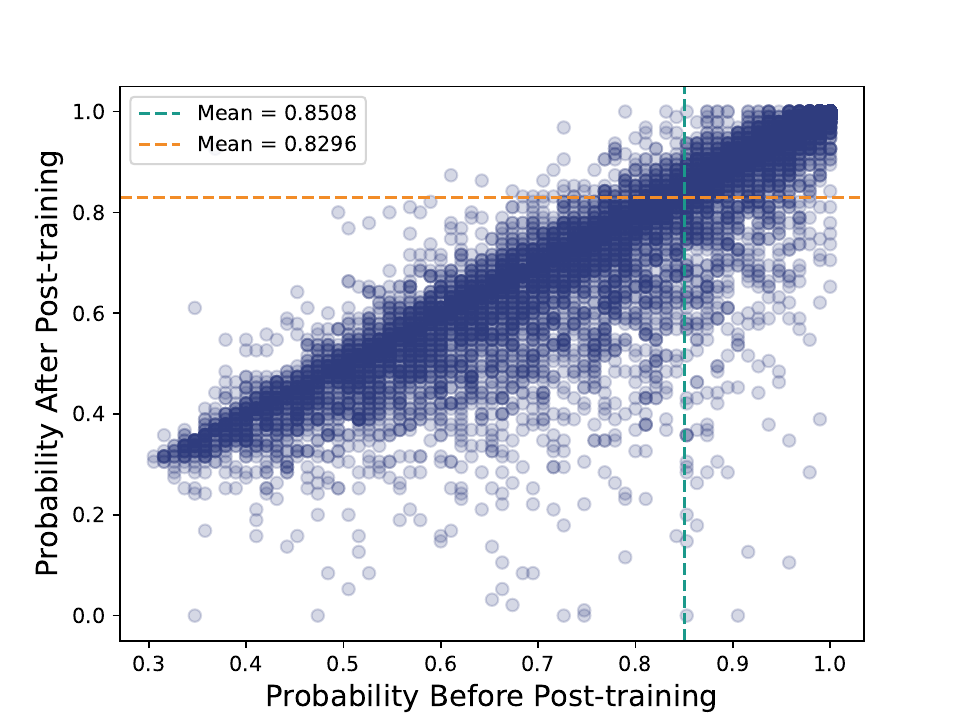}
\caption{
\textbf{Feature reappearance probabilities before and after post-training.}
The most-probable construction initializes from relatively high-probability pooled features, but post-training shifts some features toward lower reappearance probabilities. Lower-stability directions can therefore be reconstruction-useful rather than mere random artifacts.
}
\label{fig:probs_before_after}
\end{figure}

\textbf{Constructing SAEs from feature pools.}
A third route is to aggregate reproducible structure across runs rather than changing the single-run objective. Our most-probable feature-pool construction deduplicates decoder features across independently trained SAEs, selects high-probability features, and briefly post-trains the resulting dictionary. As shown in Figure~\ref{fig:ev_unstable_tradeoff}, this approach can substantially reduce instability while preserving high EV. Interestingly, before post-training the selected features all have reappearance probability bounded away from zero (around \(0.3\) in the shown run), but after post-training some features shift toward lower reappearance probabilities (Figure~\ref{fig:probs_before_after}). This supports the view that lower-stability directions can be useful for reconstruction, rather than being mere random artifacts.

\section{Automatic Interpretation and Qualitative Feature Analysis}
\label{app:qual-details}

This appendix contains the detailed qualitative results summarized in the token-entropy discussion of
Section~\ref{sec:quant}. We compare stable and unstable features using automatic feature interpretation and explanation
text analysis.

\subsection{Qualitative results from automatic interpretation}
\label{app:qual-results}

We use the same stable/unstable split as elsewhere in the paper (with $\varepsilon=0.05$ and a single anchor SAE).
We run the SAEBench auto-interpretation pipeline \citep{karvonen2025saebench0} with \texttt{Qwen/Qwen3-32B} as the
interpreter/evaluator model; full settings are given below in Appendix~\ref{app:autointerp-setup}.

\begin{figure}[t]
\vspace{-1mm}
\centering
\includegraphics[width=\linewidth]{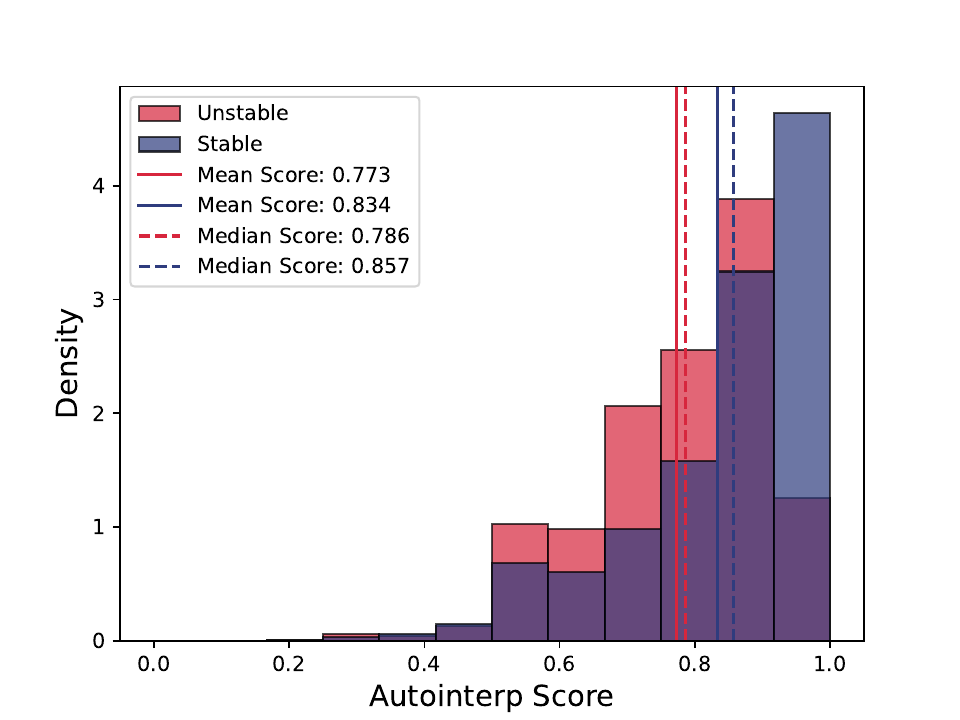}
\vspace{-2mm}
\caption{\textbf{Automatic interpretation (detection) scores for stable vs.\ unstable features.} Stable features
achieve higher scores, with a substantially larger mass at score $=1$.}
\label{fig:autointerp}
\vspace{-2mm}
\end{figure}

\textbf{Stable features are more interpretable.}
SAEBench assigns each feature a \emph{detection score}, defined as the evaluator model's accuracy at predicting whether a
given context contains \emph{any} token position on which the feature activates (Appendix~\ref{app:detection-score}).
Figure~\ref{fig:autointerp} shows the distribution of detection scores for stable and unstable features.
Stable features achieve noticeably higher scores overall.
This gap is driven primarily by the mass at score $=1$ (perfect detection on held-out contexts):
in our setting, the fraction of stable features with score $1$ is $\approx 4.5\times$ larger than for unstable features.

\textbf{Explanations differ systematically.}
Inspecting the generated explanations reveals a consistent qualitative split.
Unstable features are frequently described in terms of \emph{surface form}: letter patterns, capitalization, prefixes,
punctuation, and short substrings.
Stable features more often describe \emph{how words function in context}: phrases and constructions, syntactic roles
(e.g., pronouns, conjunctions), or broader semantic groupings.
This aligns with Section~\ref{sec:quant}, where unstable features concentrate on lower token-entropy regimes dominated by
punctuation/subword triggers, while stable features extend to higher-entropy lexical and conceptual patterns.

Table~\ref{tab:interp-examples} provides representative examples illustrating this surface-form vs. structural distinction.

\textbf{Keyword evidence for surface vs.\ structural patterns.}
To quantify the surface-vs-structural distinction in explanation text, we measure how often explanations contain
keywords such as \texttt{substring}, \texttt{starting}, \texttt{letter} (surface-form) versus \texttt{phrase},
\texttt{expression}, \texttt{construction} (structural/compositional), across bins of $\hat{p}(\boldsymbol{f}^{(0)}_i)$.
We restrict to explanations with detection score $>0.7$ to reduce noise from low-quality explanations.
A simple summary already shows a large separation: in the unstable features' explanations, \texttt{substring} appears in
$38.9\%$ of explanations, while for the stable ones it appears in $11.3\%$; conversely,
\texttt{phrase} increases from $4.1\%$ to $32.0\%$ from unstable to stable features.
The full table (with confidence intervals and bin counts) is given in Appendix~\ref{app:keyword-analysis}.

\textbf{Explanations alone predict stability.}
As an auxiliary test, we ask a frontier LLM to classify features as stable vs.\ unstable using \emph{only} their
auto-interp explanations, with labels anonymized during prompting.
On a balanced evaluation set, the model achieves $0.88$ accuracy.
Full prompts and the exact protocol are given in Appendix~\ref{app:gpt5-prompts}.

\begin{table*}[t]
\centering
\scriptsize
\begin{tabular}{p{0.48\textwidth}p{0.48\textwidth}}
\toprule
\textbf{Unstable features} & \textbf{Stable features} \\
\midrule
\textit{words starting with ``ve'' followed by a letter that continues a root or morpheme}
&
\textit{the phrase ``so as to'' and similar constructions indicating purpose or intent}
\\ \addlinespace[0.6ex]
\textit{proper nouns, especially those starting with capital letters}
&
\textit{reflexive pronouns and words referring to oneself or others}
\\ \addlinespace[0.6ex]
\textit{the substring ``Tor'' at the start of words}
&
\textit{technical terms related to servers, connections, and computing components}
\\
\addlinespace[0.6ex]
\textit{words starting with ``Div'' or containing the substring ``div'' in various contexts} &
\textit{phrases involving positional prepositions like ``in the,'' ``right up to,'' and ``next to.''} \\
\addlinespace[0.6ex]
\textit{the substring ``wr'' at the start of words} &
\textit{phrases indicating reference to previously mentioned information, such as ``as described above'' or ``mentioned earlier''} \\
\addlinespace[0.6ex]
\textit{the substring ``Tro'' at the start of words} &
\textit{phrases describing characters and their roles in stories or films} \\
\addlinespace[0.6ex]
\textit{words starting with the substrings ``wa'', ``por'', ``po'', or similar letter combinations} &
\textit{words related to peace and calmness, including variations like ``peaceable'' and ``peacefully''} \\
\addlinespace[0.6ex]
\textit{partial words with double letters in the middle, such as ``Perr'', ``Phys'', ``Ol'', ``Bonni'', ``feat'', ``iodin'', and ``Coleman''} &
\textit{the word ``firm'' and similar capitalized terms in business or organizational contexts} \\
\addlinespace[0.6ex]
\textit{words containing the substring ``cran'' followed by a letter, often related to ``cranberries'' or ``cranium.''} &
\textit{words starting with the letter ``L'' (case-insensitive)} \\
\addlinespace[0.6ex]
\textit{first names, especially when preceded by a description or title} &
\textit{words ending in the substring ``-el''} \\
\addlinespace[0.6ex]
\textit{words starting with ``Rum'', ``mist'', and ``of'' as well as variations like ``Rumor'', ``misten'', and ``of''} &
\textit{the word ``lock'' and variations with attached punctuation or nearby words} \\
\addlinespace[0.6ex]
\textit{words containing the substring ``br'' followed by ``on'' or ``o'' in medical or anatomical contexts} &
\textit{the phrase ``no longer'' and related negated continuations} \\
\addlinespace[0.6ex]
\textit{words starting with ``El'' or ``el''} &
\textit{names and parts of names starting with ``Mc'' or ``McK'' or ``Mac''} \\
\addlinespace[0.6ex]
\textit{surnames and partial surnames, especially those starting with ``Hart''} &
\textit{words related to negative outcomes or effects, especially when paired with terms like ``impact,'' ``effects,'' or ``feedback.''} \\
\addlinespace[0.6ex]
\textit{the word ``batch'' and similar capitalized substrings like ``Dow'' and ``Alabama''} &
\textit{words indicating uncertainty or hypothetical situations, such as ``potential,'' ``possible,'' and similar variations} \\
\addlinespace[0.6ex]
\textit{names or parts of names with a specific stylistic or phonetic pattern, particularly those beginning with \texttt{\textless\textless Alfred\textgreater\textgreater}, \texttt{\textless\textless styl\textgreater\textgreater}, \texttt{\textless\textless transc\textgreater\textgreater}, or \texttt{\textless\textless University\textgreater\textgreater}} &
\textit{phrases involving change with quantity or degree, especially those using ``the\ldots the\ldots'' or ``increases/decreases with.''} \\
\addlinespace[0.6ex]
\textit{words related to storage containers, especially ``tank,'' ``tanks,'' and variations like ``gallons,'' ``empty,'' and ``tankers.''} &
\textit{the word ``healthy'' and related concepts about diet and well-being} \\
\addlinespace[0.6ex]
\textit{words starting with ``pie'' or ``piez'' and related technical or compound terms} &
\textit{variations of the word ``dig'' and related actions or contexts} \\
\addlinespace[0.6ex]
\textit{the surname ``Gordon'' and related proper nouns} &
\textit{adjectives and phrases starting with ``un-'' that convey a sense of freedom or lack of restriction} \\
\addlinespace[0.6ex]
\textit{words starting with the substring ``Whit'' or ``Wal''} &
\textit{coordinating conjunctions ``and'' and ``or'' and their surrounding context} \\
\addlinespace[0.6ex]
\textit{government-related terms such as ``Cabinet,'' ``Prime,'' and ``bills.''} &
\textit{words related to teenagers and specific substrings like ``teen'', ``teens'', ``Teen'', and ``agers''} \\
\addlinespace[0.6ex]
\textit{words related to substances, fluids, or materials in contexts of waste, water, or bodily fluids} &
\textit{the word ``course'' and related educational concepts} \\
\addlinespace[0.6ex]
\textit{capitalized names and the substring ``Marshall'' with variations} &
\textit{language names and related terms such as English, Spanish, Arabic, and their variations} \\
\bottomrule
\end{tabular}
\vspace{-2mm}
\caption{
\textbf{Representative automatic interpretations.}
Examples of unstable surface-form explanations and stable structural/compositional explanations. Only interpretations with detection score $>0.7$ are shown.}
\label{tab:interp-examples}
\end{table*}

\subsection{Auto-interpretation SAEBench setup}
\label{app:autointerp-setup}

We use the SAEBench auto-interpretation pipeline \citep{karvonen2025saebench0} with \texttt{Qwen/Qwen3-32B} as both interpreter and
evaluator model.
Interpretation/evaluation contexts are drawn from a held-out slice of FineWeb consisting of $2$M tokens.
For each feature, we build prompt contexts from token windows around activations and cap the model's generation length at
$128$ tokens.

\textbf{Example selection.}
For each feature, we construct two disjoint sets of contexts:
\begin{itemize}
\item \textbf{Explanation set (used to generate the explanation):}
10 contexts with the largest activation value, plus 5 additional contexts sampled via importance sampling.
\item \textbf{Evaluation set (used to score the explanation):}
2 additional top-activation contexts (not overlapping with the explanation set), plus 10 importance-sampled contexts, plus
2 random contexts.
\end{itemize}
All sampling is performed independently per feature. The resulting prompts follow SAEBench defaults, aside from the local
inference adjustments described above.

\subsection{Detection score definition}
\label{app:detection-score}

Given a feature $i$ and a context window $c$ (a sequence of tokens), define the binary label
\[
y_i(c)=\mathbf{1}\{\exists\ \text{a token position in }c\text{ such that }z_{t,i}>0\},
\]
i.e., whether the context contains \emph{any} position where the feature activates.
Given the interpreter-produced explanation text $\mathcal{E}_i$, the evaluator model predicts $\hat{y}_i(c)\in\{0,1\}$.
The \emph{detection score} for feature $i$ is the accuracy on the held-out evaluation set:
\[
\mathrm{DS}_i \;=\; \frac{1}{|\mathcal{C}_i|}\sum_{c\in\mathcal{C}_i}\mathbf{1}\{\hat{y}_i(c)=y_i(c)\},
\]
where $\mathcal{C}_i$ is the evaluation context set described in Appendix~\ref{app:autointerp-setup}.

\subsection{Keyword analysis across stability bins}
\label{app:keyword-analysis}

We bin features by their empirical reappearance rate $\hat{p}(\boldsymbol{f}^{(0)}_i)$ and compute, for each bin and each
keyword $k$, the empirical fraction $\hat{q}$ of explanations that contain $k$ as a substring match (case-sensitive, using
the raw explanation text).
To reduce noise from low-quality explanations, we restrict to features with detection score $\mathrm{DS}_i>0.7$.

We report confidence intervals for each proportion.
Below we include two versions of the same table: a Wald interval (symmetric; matches the common $\hat{q}\pm 1.96\sqrt{\hat{q}(1-\hat{q})/n}$ form)
and a Wilson interval (typically better calibrated near 0/1). Table~\ref{tab:keyword-frac} reports the resulting keyword
frequencies with Wald confidence intervals.

\begin{table*}[t]
\centering
\small
\begin{tabular}{lcccc}
\toprule
Keyword & $[0.00,0.05)$ ($n{=}1020$) & $[0.05,0.10)$ ($n{=}220$) & $[0.90,0.95)$ ($n{=}1053$) & $[0.95,1.00]$ ($n{=}6547$) \\
\midrule
\texttt{substring}     & 0.389$\pm$0.030 & 0.209$\pm$0.054 & 0.121$\pm$0.020 & 0.113$\pm$0.008 \\
\texttt{starting}      & 0.276$\pm$0.027 & 0.082$\pm$0.036 & 0.050$\pm$0.013 & 0.052$\pm$0.005 \\
\texttt{letter}        & 0.108$\pm$0.019 & 0.068$\pm$0.033 & 0.036$\pm$0.011 & 0.037$\pm$0.005 \\
\texttt{phrase}        & 0.041$\pm$0.012 & 0.205$\pm$0.053 & 0.313$\pm$0.028 & 0.320$\pm$0.011 \\
\texttt{expression}    & 0.010$\pm$0.006 & 0.036$\pm$0.025 & 0.056$\pm$0.014 & 0.058$\pm$0.006 \\
\texttt{construction}  & 0.008$\pm$0.005 & 0.032$\pm$0.023 & 0.066$\pm$0.015 & 0.060$\pm$0.006 \\
\bottomrule
\end{tabular}
\vspace{-2mm}
\caption{
\textbf{Keyword frequencies in feature explanations across reappearance bins (Wald 95\% CI).}
Each entry reports the fraction of explanations containing the given keyword, shown as $\hat{q}\pm$ (95\% CI).
Column headers give bin ranges for $\hat{p}(\boldsymbol{f}^{(0)}_i)$ and the number of features $n$ in that bin after
restricting to detection score $>0.7$.
}
\label{tab:keyword-frac}
\end{table*}

\subsection{LLM-only classification from explanation text}
\label{app:gpt5-prompts}

We test whether the \emph{auto-interp explanation text alone} contains enough signal to predict feature stability.
Concretely, we run a blind classification experiment where a frontier LLM (GPT-5) receives only short natural-language
explanations produced by the auto-interpretation pipeline summarized in Appendix~\ref{app:qual-details} and must predict a binary label.
No activation traces, token examples, decoder vectors, or $\hat p$ values are shown.

\textbf{Data and split.}
We construct a balanced dataset of explanations from endpoint features with $\varepsilon=0.05$ using the same anchor SAE
as in the main text.
We sample $50$ explanations from each class for in-context training (100 total) and a disjoint $50$ explanations from each
class for evaluation (100 total).
During prompting, labels are anonymized as $\{0,1\}$ and the model is not told what the labels correspond to.

\textbf{Classification prompt.}
We use a single in-context learning prompt: first provide labeled training examples (balanced across labels), then ask for
labels on the test set, requesting one label per line and no additional text.
We report accuracy on the balanced test set.

\begin{PromptBox}{Prompt A: LLM-only classification template}
You will see short explanations of patterns that activate a feature.
Each explanation has a label 0 or 1. Learn what distinguishes the labels.

TRAINING EXAMPLES:
Label: 0  Explanation: <EXPLANATION_1>
Label: 0  Explanation: <EXPLANATION_2>
...
Label: 1  Explanation: <EXPLANATION_51>
Label: 1  Explanation: <EXPLANATION_52>
...

Now predict labels for the TEST explanations.
Output one label (0 or 1) per line, in order, with no extra text.

TEST EXPLANATIONS:
1) <TEST_EXPLANATION_1>
2) <TEST_EXPLANATION_2>
...
\end{PromptBox}

\textbf{Result.}
On the balanced test set (50 explanations per class), GPT-5 achieves accuracy $0.88$.
Since the model sees only explanation text, this indicates that stability is strongly reflected in the linguistic
descriptions produced by auto-interpretation.

\begin{PromptBox}{Prompt B: Qualitative class-summary template}
Here are explanations from two classes (A and B). Summarize typical patterns in each class.
Be concise and concrete.

Class A:
- <EXPLANATION_A1>
- <EXPLANATION_A2>
...

Class B:
- <EXPLANATION_B1>
- <EXPLANATION_B2>
...
\end{PromptBox}

\textbf{Qualitative class summaries prompt.}
In addition to classification, we ask the same LLM to summarize typical patterns in each class, using explanations only.
We provide a small random subset from each class (without mentioning stability) and request concise descriptions.

\textbf{Model-produced summaries.}
The model consistently describes the two classes in terms of surface-form versus structural/compositional language.
In a representative run it summarized:
\begin{itemize}
\item \textbf{Unstable:} explanations emphasize what the word looks like on the surface (substrings, prefixes,
capitalization, punctuation) and narrow lexical triggers.
\item \textbf{Stable:} explanations emphasize how words function in sentences (phrases, constructions, discourse roles)
and broader compositional or semantic groupings.
\end{itemize}

\textbf{Notes and limitations.}
This experiment is intended as a lightweight diagnostic rather than a primary result.
Because it relies on a particular evaluator model and prompt format, we treat it as supportive evidence that the
surface-vs-structural distinction is present in explanation text, consistent with Table~\ref{tab:interp-examples} and the
keyword analysis in Appendix~\ref{app:keyword-analysis}.

\section{Additional Details for Geometric Analysis}
\label{app:geometric-details}

\begin{figure}[t]
\vspace{-1mm}
\centering
\includegraphics[width=0.85\linewidth]{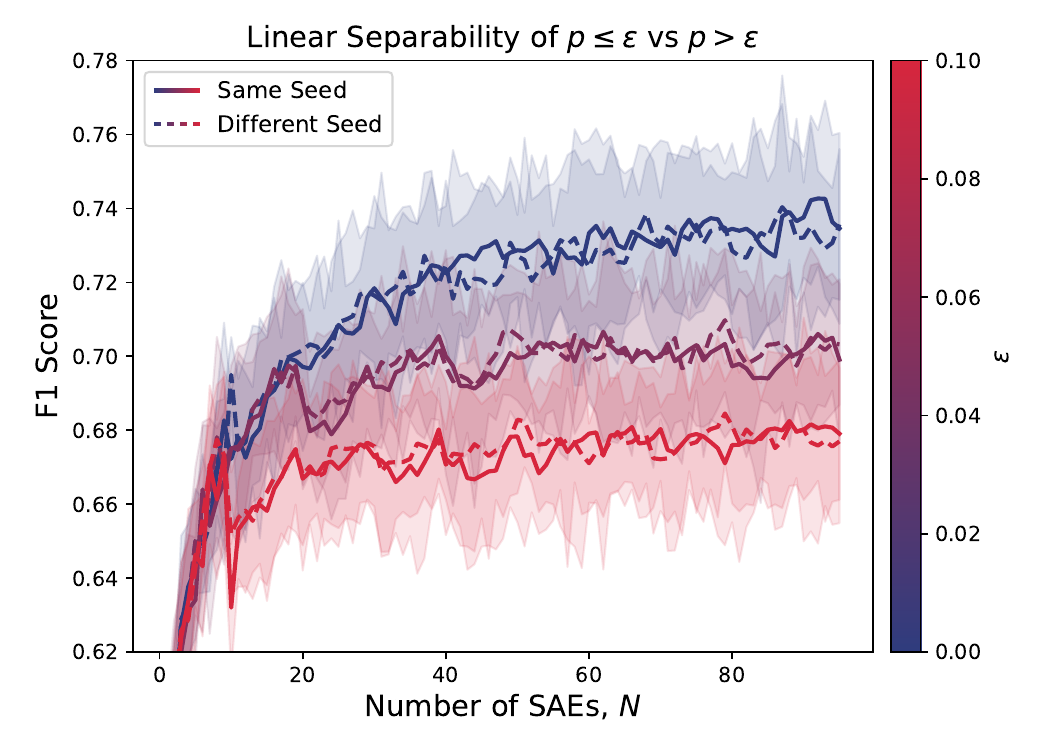}
\vspace{-2mm}
\caption{
\textbf{F1 score for classification separating unstable features from all others.} Solid curves report within-seed evaluation, while dashed curves report transfer of the classifier to a different seed.
}
\label{fig:f1_per_eps_neighbor}
\vspace{-2mm}
\end{figure}

Figure~\ref{fig:f1_per_eps_neighbor} provides the full F1 curves for the linear-separability classifier discussed in the
main text.

\subsection{Effective Rank and SVD of the Decoder Submatrices} \label{sec:svd}
As in the main text, given an index set \(\mathcal{I}^{(s)}_{\varepsilon}\subseteq\{1,\dots,F\}\) (either
\(\mathcal{U}^{(s)}_{\varepsilon}\) for unstable features or \(\mathcal{S}^{(s)}_{\varepsilon}\)
for stable features), define the corresponding submatrix
\[
X^{(s)}_{\mathcal{I}_\varepsilon}
\;:=\;
\boldsymbol{W}^{(s)}_{\text{dec}}[:, \mathcal{I}^{(s)}_{\varepsilon}]
\in
\mathbb{R}^{d\times m_s},
\qquad
m_s:=|\mathcal{I}^{(s)}_{\varepsilon}|.
\]

We consider its singular value decomposition
\begin{equation*}
X^{(s)}_{\mathcal{I}_\varepsilon}
\;=\;
V^{(s)}_{\varepsilon}\,\Sigma^{(s)}_{\varepsilon}\,\big(U^{(s)}_{\varepsilon}\big)^\top,
\end{equation*}
where
\(V^{(s)}_{\varepsilon}\in\mathbb{R}^{d\times d}\)
and
\(U^{(s)}_{\varepsilon}\in\mathbb{R}^{m_s\times m_s}\)
are orthogonal matrices, and
\(\Sigma^{(s)}_{\varepsilon}\in\mathbb{R}^{d\times m_s}\)
is rectangular, with diagonal entries
\(\sigma^{(s)}_1\ge \sigma^{(s)}_2\ge \cdots \ge \sigma^{(s)}_{\text{min}(d,m_s)}\ge 0\), that are singular values,
and all remaining diagonal entries are zero.

We define the effective rank of submatrix $X^{(s)}_{\mathcal{I}_\varepsilon}$ as
\[
\mathrm{ER}(X^{(s)}_{\mathcal{I}_\varepsilon})
=
\exp\!\Big(-\sum_{k=1}^{\text{min}(d, m_s)}p^{(s)}_k\,\log p^{(s)}_k\Big),
\]
where $p^{(s)}_k=\sigma^{(s)}_k/{\sum_{j=1}^{\text{min}(d, m_s)} \sigma^{(s)}_j}$. To fairly compare the effective ranks of submatrices with different numbers of features, we calculate them under a size-matched protocol: for each seed for all feature sets we select
\[
k^{(s)} \;=\; \min\big(|\mathcal{U}^{(s)}_{0.01}|,\;|\mathcal{S}^{(s)}_{0.01}|, |\mathcal{U}^{(s)}_{0.05}|,\;|\mathcal{S}^{(s)}_{0.05}|\big),
\]
and for sets whose number of features differs from $k^{(s)}$, we estimate $\mathrm{ER}$ by averaging over $B=50$ random subsamples of size $k^{(s)}$.

\begin{table}[t]
\centering
\small
\setlength{\tabcolsep}{8pt}
\begin{tabular}{lc}
\toprule
Probability Set & Effective Rank$/d$ \\
\midrule
$p \leq 0.01$ & $0.587 \pm 0.006$ \\
$p \leq 0.05$ & $0.649 \pm 0.006$ \\
$p \geq 0.95$ & $0.808 \pm 0.004$ \\
$p \geq 0.99$ & $0.804 \pm 0.004$ \\
\bottomrule
\end{tabular}
\caption{Effective rank normalized by the hidden-state dimension $d=768$, for size-matched probability sets.}
\label{tab:effective-rank-prob-sets}
\end{table}

We define the explained variance $\mathrm{EV}^{(s)}_{\text{SVD}}(r)$ of singular values of matrix $X^{(s)}_{\mathcal{I}_\varepsilon}$ as
\begin{equation*}
\mathrm{EV}^{(s)}_{\text{SVD}}(r)
=
\sum_{i=1}^{r} \big(\sigma^{(s)}_i\big)^2/\sum_{i=1}^{\text{min}(d,m_s)} \big(\sigma^{(s)}_i\big)^2.
\end{equation*}

Let \(V^{(s)}_{\varepsilon}[:,1\!:\!r]\in\mathbb{R}^{d\times r}\) denotes the matrix, whose \(r\) columns are the top-\(r\) left singular vectors, and define the projector onto their span
\[
P^{(s)}_{\varepsilon}(r)
\;:=\;
V^{(s)}_{\varepsilon}[:,1\!:\!r]\,(V^{(s)}_{\varepsilon}[:,1\!:\!r])^\top.
\]
Then \(\mathrm{EV}^{(s)}_{\text{SVD}}(r)\) admits an alternative representation
\[
\mathrm{EV}^{(s)}_{\text{SVD}}(r)
\;=\;
\frac{\big\|P^{(s)}_{\varepsilon}(r) X^{(s)}_{\mathcal{I}_\varepsilon}\big\|_F^2}{\big\|X^{(s)}_{\mathcal{I}_\varepsilon}\big\|_F^2}.
\]
This suggests a natural cross-seed analogue: for an anchor seed \(a\) and another seed \(s \neq a\), define
\[
\mathrm{EV}^{(a\to s)}_{\text{SVD}}(r)
\;:=\;
\frac{\big\|P^{(a)}_{\varepsilon}(r) X^{(s)}_{\mathcal{I}_\varepsilon}\big\|_F^2}{\big\|X^{(s)}_{\mathcal{I}_\varepsilon}\big\|_F^2}.
\]
Intuitively, \(\mathrm{EV}^{(a\to s)}_{\text{SVD}}(r)\) measures how well the top-\(r\) singular subspace learned in seed \(a\) explains the feature
subspace in seed \(s\).

\subsection{Additional Results for the Controlled Low-Rank Synthetic Model}
\label{app:toy-model}

\begin{figure}[t]
\centering
\begin{subfigure}[b]{\linewidth}
    \centering
    \includegraphics[width=\linewidth]{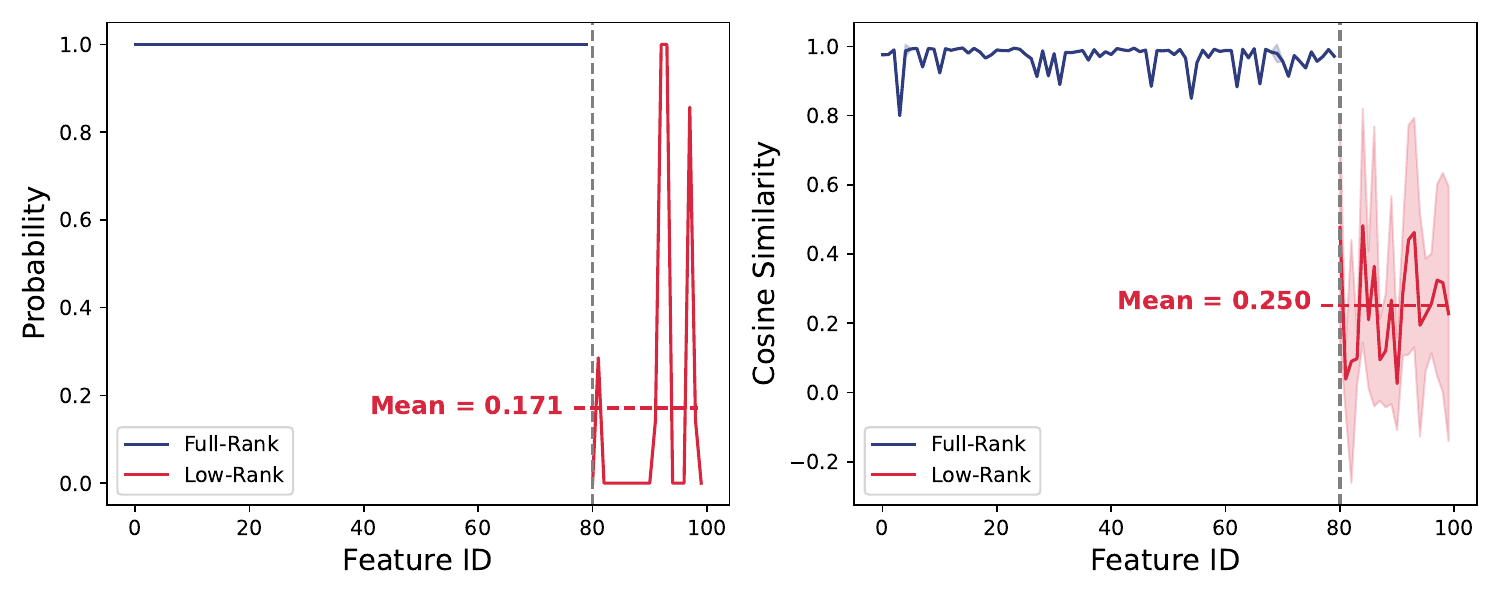}
    \caption{$d=32$, $r=1$, $k=4$.}
\end{subfigure}

\vspace{2mm}

\begin{subfigure}[b]{\linewidth}
    \centering
    \includegraphics[width=\linewidth]{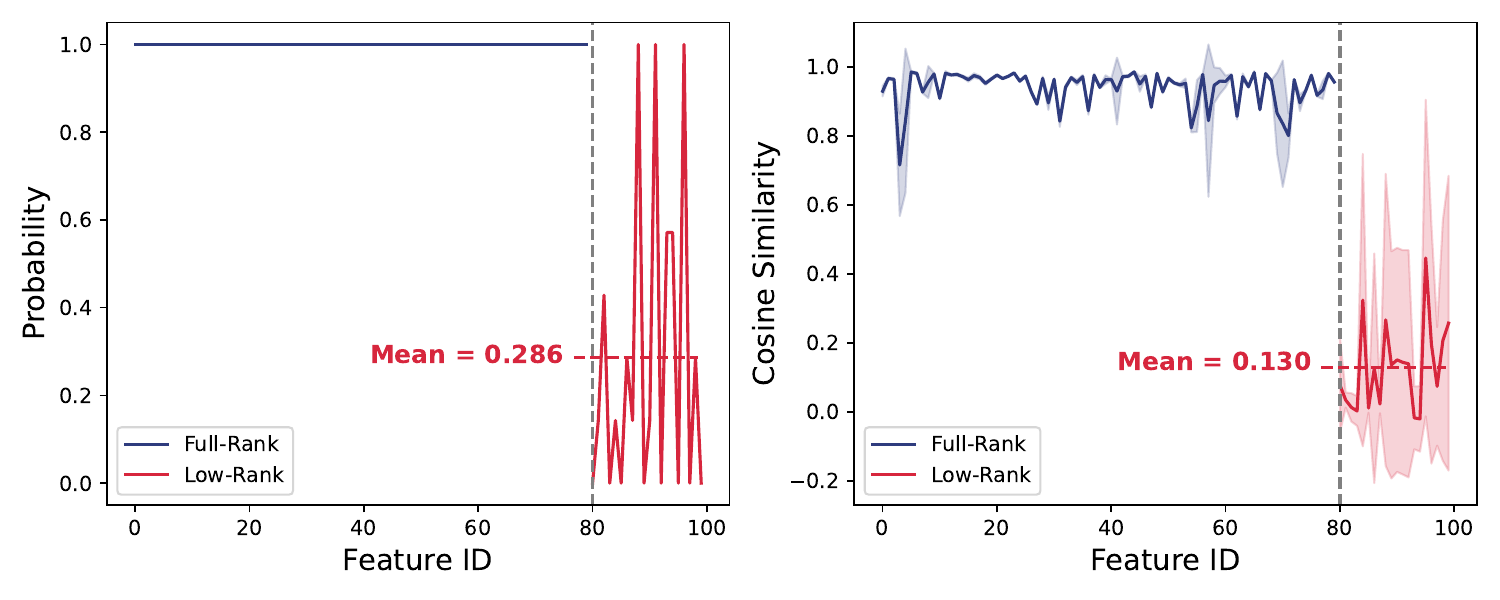}
    \caption{$d=32$, $r=1$, $k=8$.}
\end{subfigure}

\vspace{2mm}

\begin{subfigure}[b]{\linewidth}
    \centering
    \includegraphics[width=\linewidth]{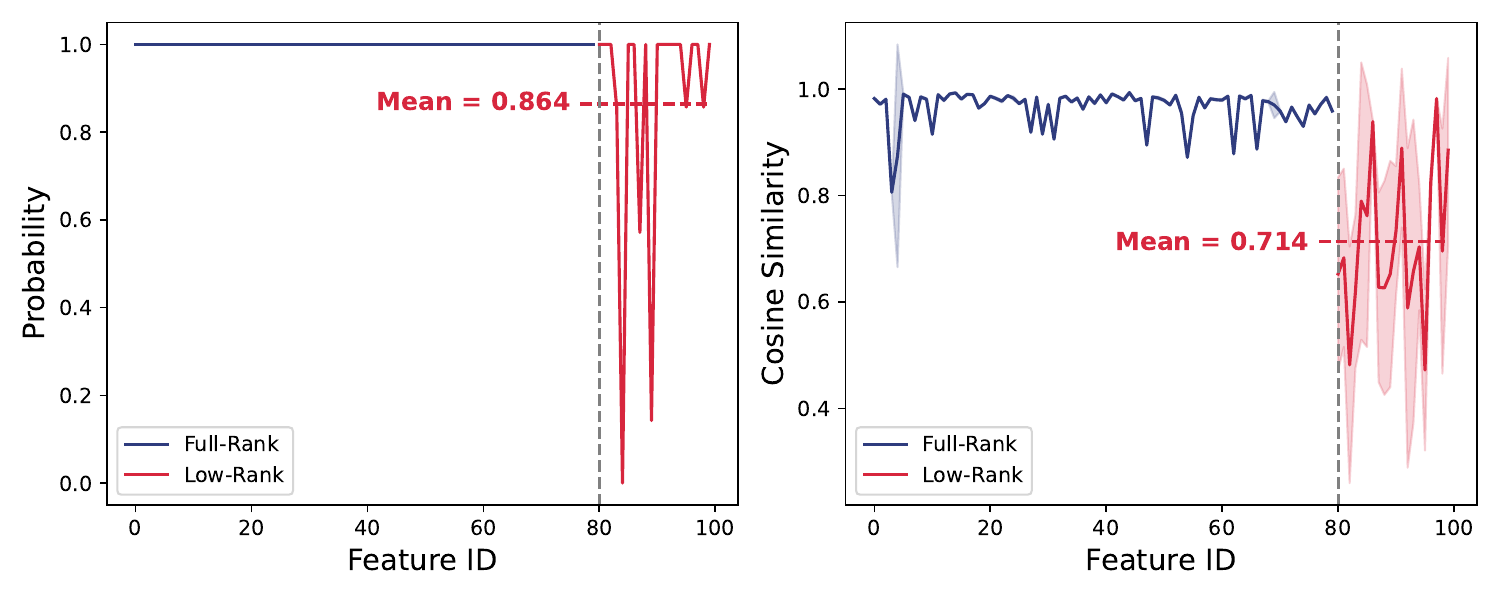}
    \caption{$d=32$, $r=4$, $k=4$.}
\end{subfigure}

\caption{
\textbf{Additional synthetic settings.}
The same full-rank versus low-rank split appears across multiple values of the subspace rank $r$ and sparsity level
$k$: full-rank features retain high reappearance probabilities and high cosine similarity to ground truth, whereas
low-rank features do not.
}
\label{fig:toy_prob_cossim_appendix}
\end{figure}

We report additional synthetic-model results beyond the main setting from Section~\ref{sec:toy-model}. Across multiple
choices of low-rank dimension $r$ and sparsity level $k$, the same qualitative pattern persists: full-rank features are
consistently stable and well recovered, while features confined to the shared low-rank block are much less stable as
individual directions. Figure~\ref{fig:toy_prob_cossim_appendix} shows these additional settings.

\begin{figure}[t]
\centering
\includegraphics[width=\linewidth]{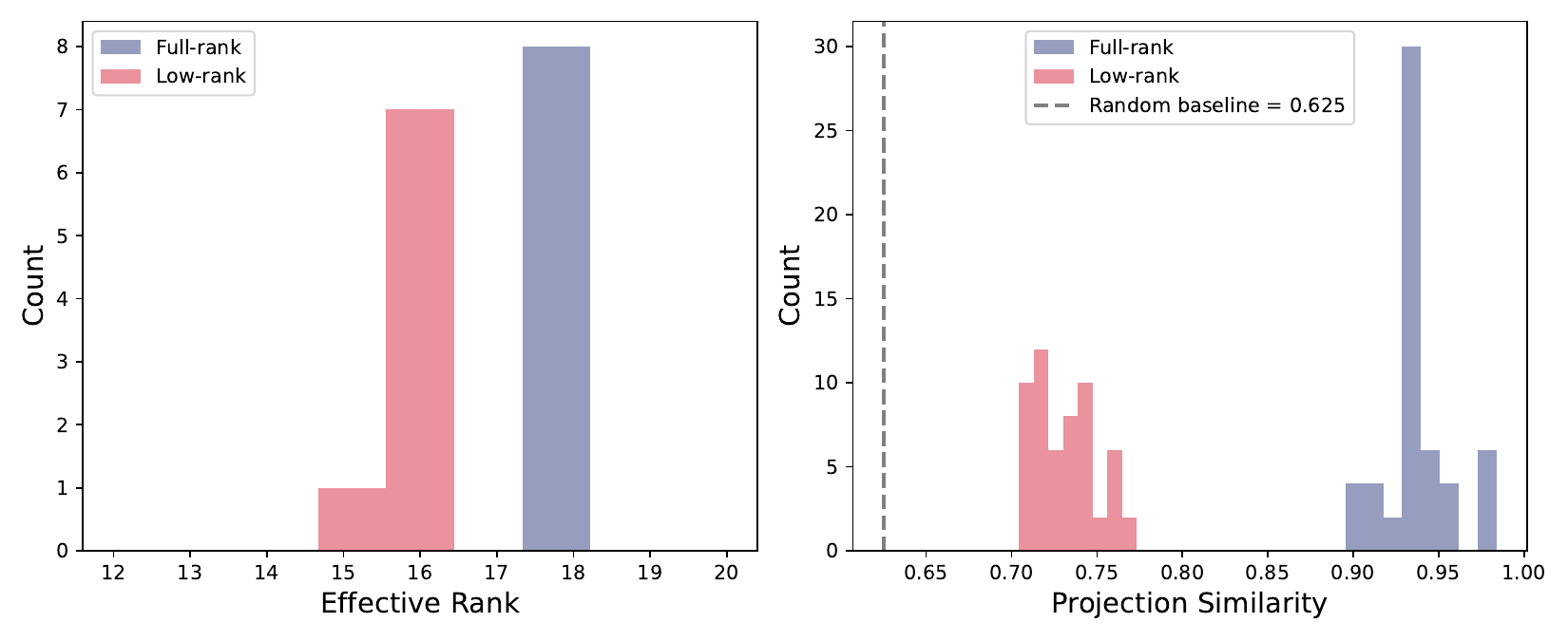}
\caption{
\textbf{Effective rank and cross-seed subspace similarity in the synthetic model ($d=32$, $r=2$, $k=8$).}
The learned low-rank block has smaller effective rank than the full-rank block, yet its cross-seed subspace similarity
remains clearly above the random baseline. Thus the low-rank features are unstable as individual vectors but still
reconstruct a reproducible shared subspace.
}
\label{fig:toy_eff_rank_subspace}
\end{figure}

Figure~\ref{fig:toy_eff_rank_subspace} reports the corresponding effective-rank and cross-seed subspace-similarity
diagnostics.

\subsubsection{Residual-Based Diagnostic for Low-Rank Feature Recovery}
\label{app:manual-identification}

The toy model also admits a diagnostic for checking whether the low-rank ground-truth atoms are present in the residual
structure, even when they are not recovered as individual SAE latents. Let an activation with exactly one low-rank
ground-truth feature be
\[
h^{(t)}
=
\sum_{i\in F_t} f_i + g_{r(t)},
\]
where \(f_i\) are full-rank ground-truth features and \(g_{r(t)}\) is a low-rank ground-truth feature. Suppose the SAE
reconstruction activates the correctly identified full-rank features and exactly one unstable feature \(\hat g_{j(t)}\):
\[
\hat h^{(t)}
\approx
\sum_{i\in F_t} f_i + \hat g_{j(t)}.
\]
Then the residual-corrected vector
\[
h^{(t)}-\hat h^{(t)}+\hat g_{j(t)}
\approx
g_{r(t)}.
\]
Let \(z^{(t)}\) denote the vector of unstable-feature activations for example \(t\). Then
\[
R
=
\left\{
h^{(t)}-\hat h^{(t)}+\hat g_{j(t)}
\;:\;
\|z^{(t)}\|_0=1
\right\}.
\]
should cluster around the ground-truth low-rank atoms \(g_1,\dots,g_M\), assuming each low-rank atom appears in at least
one selected activation. In the idealized case where the full-rank features are exactly identified and there is no
reconstruction error, the elements of \(R\) are exactly ground-truth low-rank atoms; with approximate full-rank recovery,
they are noisy samples around those atoms.

\begin{figure}[t]
\centering
\includegraphics[width=\linewidth]{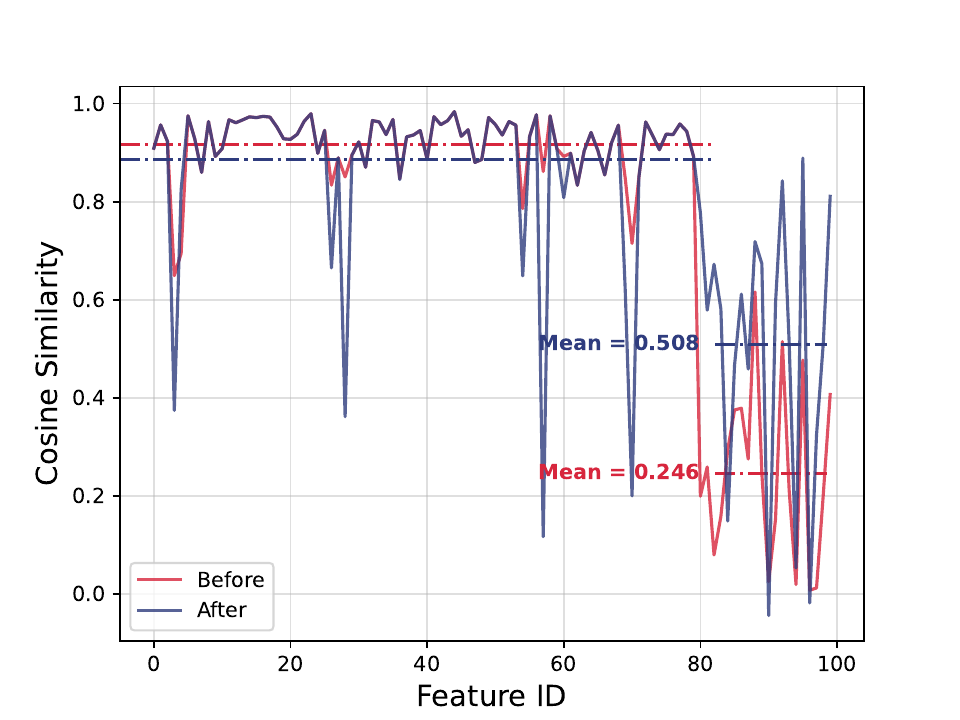}
\caption{
\textbf{Residual-based recovery of low-rank ground-truth features in the toy model.}
Cosine similarity between decoder rows and matched ground-truth features before and after replacing unstable decoder rows
with centroids obtained from residual-corrected activations. The procedure partially recovers the low-rank block while
leaving the already identified full-rank block largely unchanged.
}
\label{fig:manual_identification}
\end{figure}

We therefore run \(k\)-means on \(R\) with \(M\) clusters, replace the unstable decoder rows by the resulting centroids,
and recompute cosine similarity to the ground-truth dictionary. Figure~\ref{fig:manual_identification} shows that this
substantially improves alignment for the low-rank block in the \(d=32,r=2,k=8\) toy setting. This is a diagnostic rather
than a practical training method: it relies on the toy setup's known number of low-rank atoms and selected one-unstable
examples, and freezing the replaced decoder rows followed by encoder retraining did not improve EV.

\section{Additional Details for Other Setups}
\label{app:other-setups}

\subsection{Endpoint fractions across models, layers, and dictionary sizes}
\label{app:endpoints-topk}

\Cref{sec:other-setups} summarizes how endpoint stability varies across base models, layers, and dictionary sizes for TopK
SAEs. \Cref{tab:endpoints-topk} reports the full endpoint fractions for all configurations considered. 

\begin{figure}[!t]
\vspace{-1mm}
\centering
\includegraphics[width=1.0\linewidth]{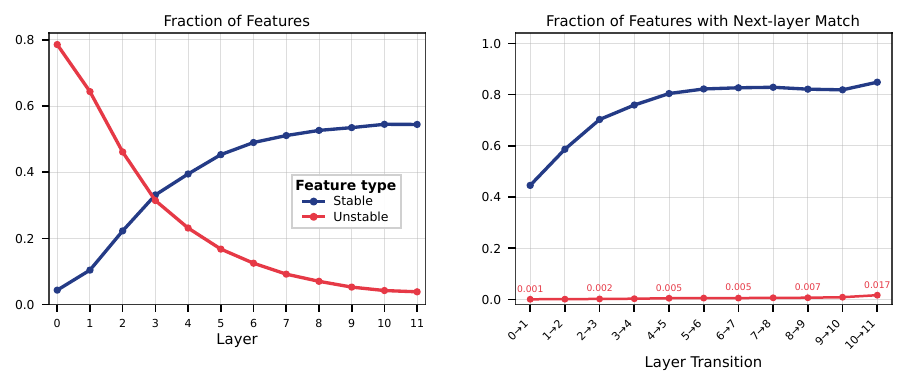}
\vspace{-2mm}
\caption{\textbf{Evolution of feature stability across layers.}
Left: layer-wise fraction of stable (blue) and unstable (red) SAE features, showing how reproducible and non-reproducible dictionary elements evolve with depth. Right: for stable (blue) and unstable (red) source features separately, the fraction with a next-layer decoder-cosine match at threshold $\tau_{\mathrm{layer}}=0.7$, measuring cross-layer feature matching.
}
\label{fig:evolution_of_stability}
\vspace{-2mm}
\end{figure}

The left panel of Figure~\ref{fig:evolution_of_stability} reports the layer-wise fractions of stable and unstable features in TopK SAEs with \(K=64\) and \(F=2^{14}\), trained on GPT--2 residual-stream activations, using 21 independently trained SAEs per layer. 
The right panel of Figure~\ref{fig:evolution_of_stability} presents the fraction of stable and unstable source features with a next-layer decoder-cosine match at threshold \(\tau_{\mathrm{layer}}=0.7\). Overall, Figure~\ref{fig:evolution_of_stability} shows that stable SAE features become more prevalent with layer depth and are much more likely than unstable features to admit a next-layer decoder-cosine match.

\begin{table*}[t]
\centering
\caption{
Fractions of \emph{unstable} ($\hat{p}\leq \varepsilon$) and \emph{stable} ($\hat{p}\geq 1-\varepsilon$) features for TopK
SAEs across dictionary sizes and layers.
}
\label{tab:endpoints-topk}
\small
\setlength{\tabcolsep}{5pt}
\begin{tabular}{ll *{3}{cc}}
\toprule
\multirow{2}{*}{Base model} & \multirow{2}{*}{Layer} &
\multicolumn{2}{c}{$F=2^{14}$} & \multicolumn{2}{c}{$F=2^{15}$} & \multicolumn{2}{c}{$F=2^{16}$} \\
\cmidrule(lr){3-4}\cmidrule(lr){5-6}\cmidrule(lr){7-8}
& & Unstable & Stable & Unstable & Stable & Unstable & Stable \\
\midrule

\multirow{3}{*}{GPT-2}
& 4  & 0.219 & 0.339 & 0.161 & 0.356 & 0.088 & 0.387 \\
& 7  & 0.080 & 0.447 & 0.066 & 0.440 & 0.051 & 0.432 \\
& 10 & 0.031 & 0.484 & 0.035 & 0.470 & 0.040 & 0.433 \\
\midrule

\multirow{3}{*}{Pythia-160M-deduped}
& 4  & 0.136 & 0.421 & 0.109 & 0.558 & 0.049 & 0.634 \\
& 7  & 0.062 & 0.513 & 0.096 & 0.569 & 0.468 & 0.316 \\
& 10 & 0.064 & 0.461 & 0.077 & 0.527 & 0.114 & 0.456 \\
\midrule

\multirow{3}{*}{Gemma-2 2B}
& 7  & 0.042 & 0.653 & 0.027 & 0.647 & 0.021 & 0.588 \\
& 12 & 0.013 & 0.563 & 0.012 & 0.548 & 0.024 & 0.482 \\
& 18 & 0.005 & 0.681 & 0.008 & 0.629 & 0.020 & 0.509 \\
\bottomrule
\end{tabular}
\end{table*}

\subsection{Training-token sweeps: setup and measurement details}
\label{app:token-sweeps}

In \Cref{sec:other-setups} we study how endpoint stability evolves as SAE training progresses.
We sweep the total number of SAE training tokens from $10$M to $10$B while keeping the SAE architecture and
hyperparameters fixed to the main TopK setting (GPT-2, \texttt{resid\_post} layer 7, TopK=$64$, $F=2^{14}$).
For each training budget, we train the same number of independent SAEs (different random seeds) and estimate
feature reappearance probabilities using the matching procedure in \Cref{sec:stability-metric}.
We then report the endpoint fractions (unstable and stable) using the same $\varepsilon$ and cosine threshold
$\theta$ as in the main setting.

\begin{figure}[t]
\vspace{-1mm}
\centering
\includegraphics[width=\linewidth]{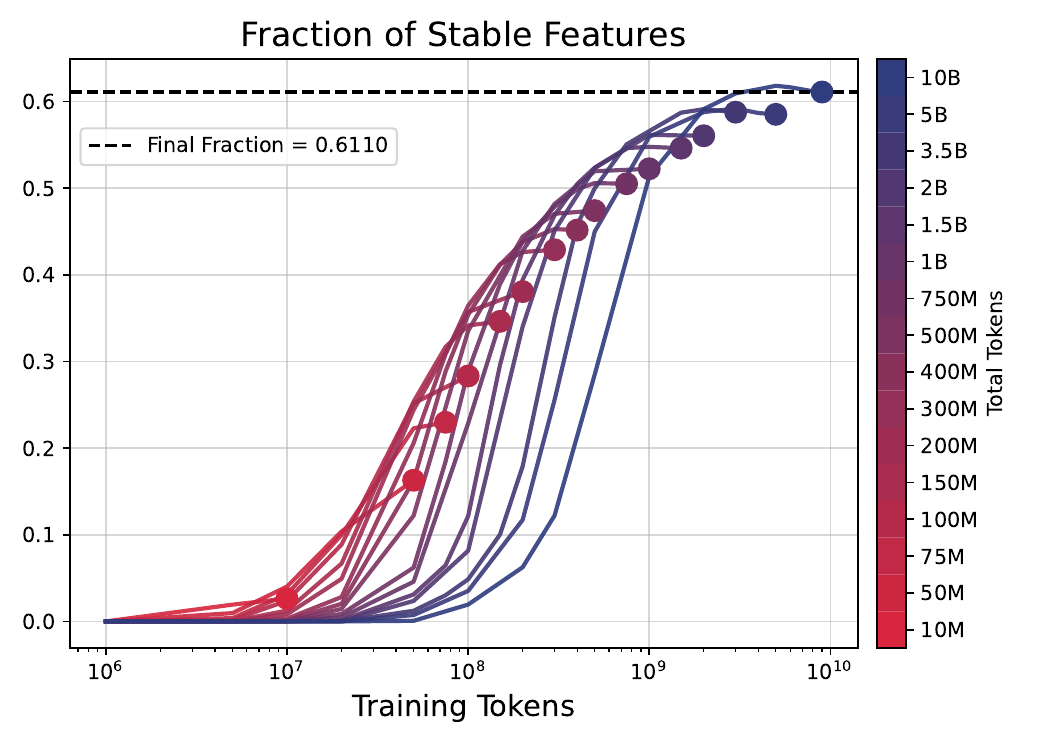}
\vspace{-2mm}
\caption{\textbf{Stable fraction vs.\ SAE training tokens.} Fraction of stable features in the main TopK setting as a
function of total SAE training tokens.}
\label{fig:stable_vs_tokens}
\vspace{-2mm}
\end{figure}

Figure~\ref{fig:unstable_vs_tokens} reports the fraction of unstable features as a function of training tokens.
Figure~\ref{fig:stable_vs_tokens} reports the corresponding stable fraction curve.
Together, these curves show that instability decreases early and then approaches a non-zero plateau, while the
stable fraction continues to increase with diminishing returns.

\subsection{Automatic interpretation on trained vs.\ random transformers}
\label{app:autointerp-random}

\begin{figure}[t]
\vspace{-1mm}
\centering
\includegraphics[width=\linewidth]{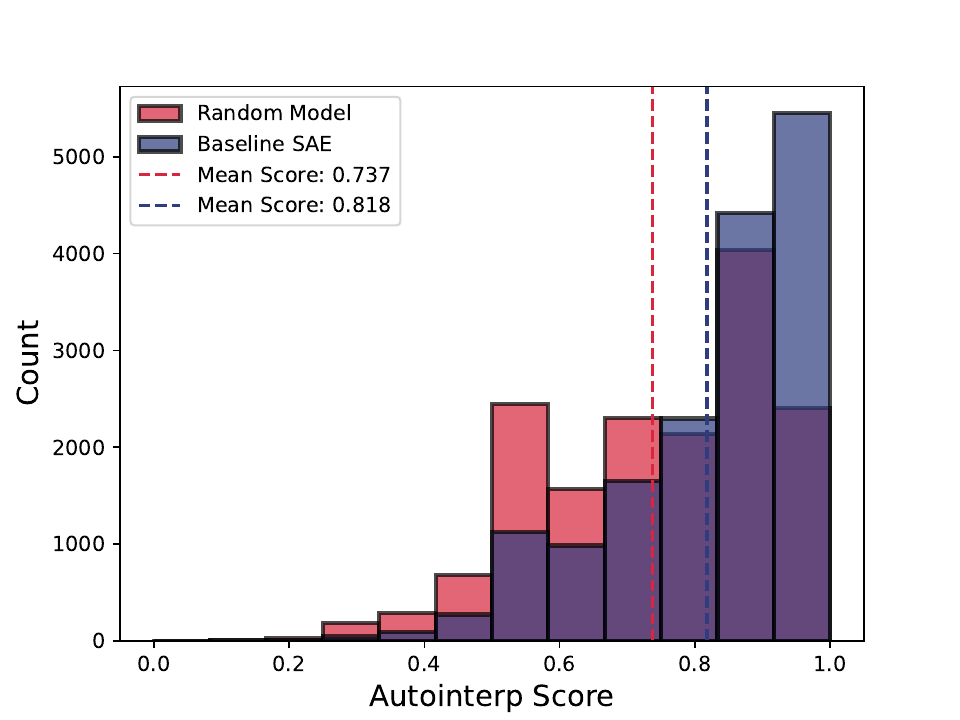}
\vspace{-2mm}
\caption{\textbf{Automatic interpretation scores for SAEs trained on trained vs.\ random base models.} High detection
scores are achievable even on random-model activations, despite the absence of cross-seed reproducibility.}
\label{fig:autointerp_random}
\vspace{-2mm}
\end{figure}

Figure~\ref{fig:autointerp_random} compares automatic interpretation (detection) scores for SAEs trained on
activations from a trained GPT-2 versus a randomly initialized GPT-2 (same SAE architecture and hyperparameters). This
shows that high detection scores can occur even in the random-model setting, motivating stability as a complementary
faithfulness signal.

\end{document}